\documentclass[10pt,twocolumn,letterpaper]{article}

\usepackage{cvpr}
\usepackage{times}
\usepackage{epsfig}
\usepackage{graphicx}
\usepackage{subfigure}
\usepackage{amsmath}
\usepackage{amssymb}
\usepackage{booktabs} 
\usepackage{multirow}
\usepackage{url}
\usepackage{enumitem}
\usepackage{microtype}
\usepackage[font=small,labelfont=bf]{caption}
\usepackage{xcolor}
\usepackage{diagbox}
\usepackage{nameref}
\usepackage{zref-xr}
\usepackage{comment}
\usepackage{nicefrac}
\zxrsetup{toltxlabel}

\usepackage[pagebackref=true,breaklinks=true,letterpaper=true,colorlinks,bookmarks=false]{hyperref}
\usepackage{cleveref}

\cvprfinalcopy % *** Uncomment this line for the final submission

\def\cvprPaperID{14} % *** Enter the CVPR Paper ID here

\def\eqnvspace{{\vspace{-2mm}}}

\newenvironment{tight_itemize}{
\begin{itemize}[leftmargin=20pt]
  \setlength{\topsep}{0pt}
  \setlength{\itemsep}{0pt}
  \setlength{\parskip}{0pt}
  \setlength{\parsep}{0pt}
}{\end{itemize}}

\newcommand{\ndim}{K}
\newcommand{\ncls}{N}
\newcommand{\dannssl}{DANN-CA}

\newcommand{\accgan}{AC-CGAN}

\newcommand{\Fig}[1]{Fig.~\ref{#1}}
\newcommand{\overbar}[1]{\mkern 2.mu\overline{\mkern-2.mu#1\mkern-2.mu}\mkern 2.mu}

\newcommand{\Paragraph}[1]{\vspace{1mm} \noindent \textbf{#1} \hspace{0mm}}
\newcommand{\Section}[1]{\vspace{-1.5mm} \section{#1} \vspace{-1.5mm}}
\newcommand{\SubSection}[1]{\vspace{-0.3mm} \subsection{#1} \vspace{-1.2mm}}

\begin{document}
\title{Gotta Adapt 'Em All: Joint Pixel and Feature-Level\\Domain Adaptation for Recognition in the Wild}

\author{Luan Tran$^1$
\thanks{This work is done when L.~Tran was an intern at NEC Labs America.}
\hspace{0.08in} Kihyuk Sohn$^{2}$
\hspace{0.08in} Xiang Yu$^2$
\hspace{0.08in} Xiaoming Liu$^1$
\hspace{0.08in} Manmohan Chandraker$^{2,3}$
\vspace{1mm} \\
\hspace{0.15in} $^{1}$Michigan State University
\hspace{0.15in} $^{2}$NEC Labs America
\hspace{0.15in} $^{3}$UC San Diego\\
}

\maketitle

% ------- abstract -------
\begin{abstract}
\vspace{-0.05in}
Recent developments in deep domain adaptation have allowed knowledge transfer from a labeled source domain to an unlabeled target domain at the level of intermediate features or input pixels. We propose that advantages may be derived by combining them, in the form of different insights that lead to a novel design and complementary properties that result in better performance. At the feature level, inspired by insights from semi-supervised learning, we propose a classification-aware domain adversarial neural network that brings target examples into more classifiable regions of source domain. Next, we posit that computer vision insights are more amenable to injection at the pixel level. In particular, we use 3D geometry and image synthesis based on a generalized appearance flow to preserve identity across pose transformations, while using an attribute-conditioned CycleGAN to translate a single source into multiple target images that differ in lower-level properties such as lighting. Besides standard UDA benchmark, we validate on a novel and apt problem of car recognition in unlabeled surveillance images using labeled images from the web, handling explicitly specified, nameable factors of variation through pixel-level and implicit, unspecified factors through feature-level adaptation.
\end{abstract}

% ------- intro -------
\begin{table}[t]
\footnotesize
\begin{center}
\includegraphics[width=0.95\linewidth]{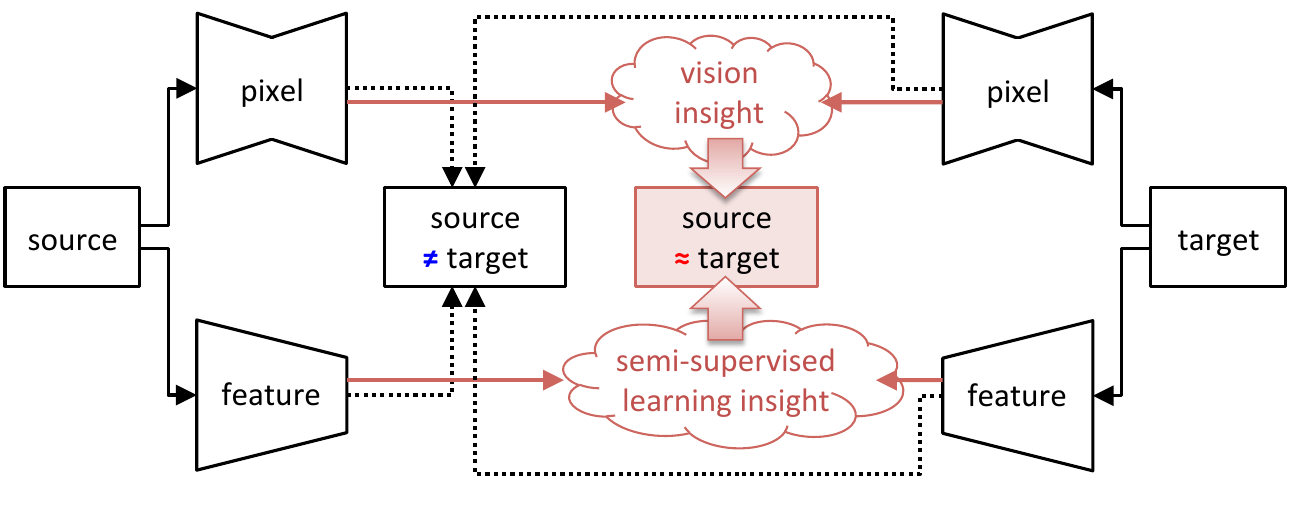}
\begin{tabular}{l|ccc}
\toprule
\multirow{2}{*}{\backslashbox{Feature}{Pixel}} & \multirow{2}{*}{--} & \multirow{2}{*}{CycleGAN} & \textbf{MKF+{\accgan}} \\
& & & \textbf{(ours)} \\
\midrule
-- & 55.0 & 64.3 & \textbf{\color{blue}79.7} \\
DANN & 60.4 & 64.8 & 78.0 \\
\textbf{{\dannssl} (ours)} & \textbf{\color{blue}75.8} & 77.7 & \textbf{\color{red}84.2} \\
\bottomrule
\end{tabular}
\end{center}
\vspace{-0.18in}
\caption{Our framework for unsupervised domain adaptation at multiple semantic levels: at \emph{feature-level}, we bring insights from semi-supervised learning to obtain highly discriminative domain-invariant representations; at \emph{pixel-level}, we leverage complementary domain-specific vision insights e.g., geometry and attributes. Our joint pixel and feature-level DA demonstrates significant improvement over individual adaptation counterparts as well as other competing methods such as CyCADA (CycleGAN+DANN)~\cite{hoffman2017cycada} on car recognition in surveillance domain under UDA setting. Please see Section~\ref{sec:exp} for complete experimental analysis.\label{fig:teasar_small}}
\vspace{-0.18in}
\end{table}

\vspace{-0.15in}
\Section{Introduction}
\label{sec:intro}
Deep learning has made an enormous impact on many applications in computer vision such as generic object recognition~\cite{krizhevsky2012imagenet,simonyan2014very,szegedy2015going,he2016deep}, fine-grained categorization~\cite{wah2011caltech,krause20133d,schroff2015facenet}, object detection \cite{Lin_etal_2017b,Lin_etal_2017a,Liu_etal_2016,Redmon_etal_2016,Ren_etal_2015}, semantic segmentation~\cite{Chen_etal_2017,Shelhamer_etal_2017} and 3D reconstruction~\cite{tran2018nonlinear, tran2019towards}. Much of its success is attributed to the availability of large-scale labeled training data~\cite{deng2009imagenet,guo2016ms}. However, this is hardly true in many practical scenarios: since annotation is expensive, most data remains unlabeled. Consider car recognition problem from surveillance images, where factors such as camera angle, distance, lighting or weather condition are different across locations. It is not feasible to exhaustively annotate all these images. Meanwhile, there exists abundant labeled data from web domain~\cite{krause20133d,yang2015large,gebru2017scalable}, but with very different image characteristics that precludes direct transfer of discriminative CNN-based classifiers. For instance, web images might be from catalog magazines with professional lighting and ground-level camera poses, while surveillance images can originate from cameras atop traffic lights with challenging lighting and weather conditions.

Unsupervised domain adaptation (UDA) is a promising tool to overcome the lack of labeled training data problem in target domains. Several approaches aim to match distributions between source and target domains at different levels of representations, such as feature \cite{tzeng2014deep,tzeng2015simultaneous,ganin2016domain,Sohn_2017_ICCV,luo_nips17_label} or pixel levels \cite{taigman2016unsupervised,shrivastava2016learning,zhu2017unpaired,Bousmalis_2017_CVPR}. Certain adaptation challenges are better handled in the feature space, but feature-level DA is a black-box algorithm for which adding domain-specific insights during adaptation is more difficult than in pixel space. On the contrary, pixel space is much higher-dimensional and the optimization problem is under-determined. How to effectively combine them has become an open challenge. 

\begin{figure}[t]
\centering
\subfigure[{DANN (baseline)}]{\includegraphics[width=0.47\textwidth]{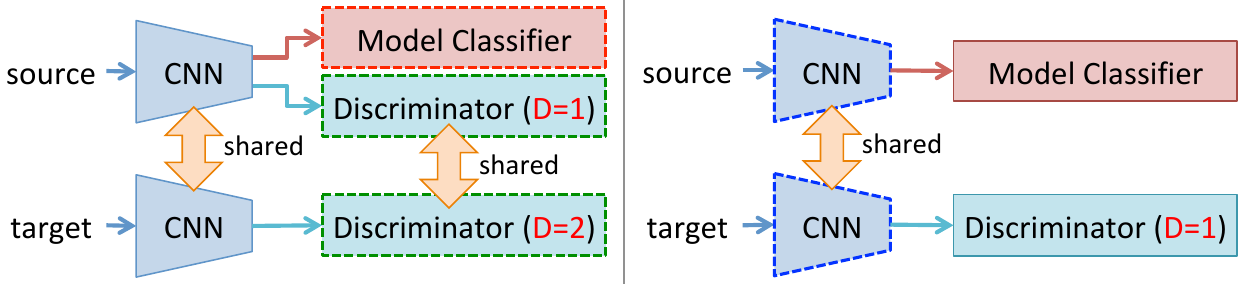}\label{fig:domadv-dann}}
\subfigure[{\dannssl}]{\includegraphics[width=0.47\textwidth]{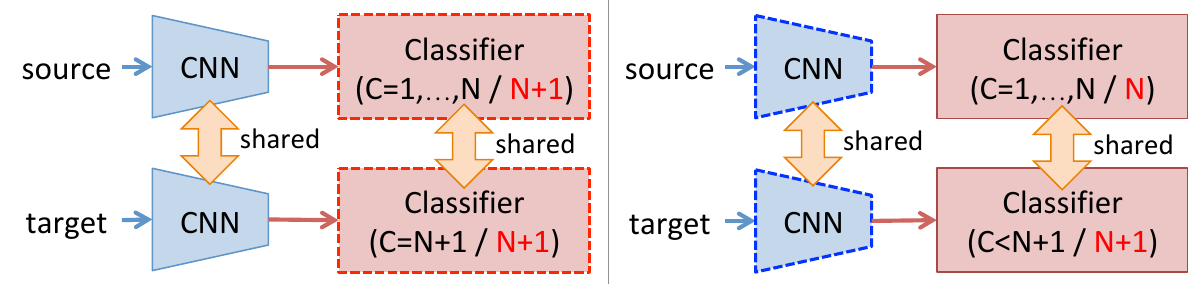}\label{fig:domadv-ss}}
\vspace{-0.1in}
\caption{(a) DANN and (b) classification-aware DANN ({\dannssl}) with $({\ncls}{+}1)$-way joint parameterization of classifier and discriminator. CNN and classifiers are updated in turn (dotted boxes) while fixing the others (solid boxes).\label{fig:domadv}}
\vspace{-0.2in}
\end{figure}

In this work we address this challenge by leveraging complementary tools that are better-suited at each level (see figure in Table~\ref{fig:teasar_small}). Specifically, we posit that feature-level DA is more amenable to techniques from semi-supervised learning (SSL), while pixel-level DA allows domain-specific insights from computer vision. In Section~\ref{sec:domadapt}, we present our feature-level DA method called classification-aware domain adversarial neural network ({\dannssl}) that jointly parameterizes the classifier and domain discriminator inspired by an instance of SSL algorithm~\cite{salimans2016improved}. We show this to be a generalization of DANN~\cite{ganin2016domain} to incorporate constraints (\Fig{fig:domadv}) that guide discriminator to easily find major modes corresponding to classes in the feature space, and in turn put target examples into more classifiable regions via adversarial loss.

\begin{figure}[t]
\centering
\includegraphics[width=0.99\linewidth]{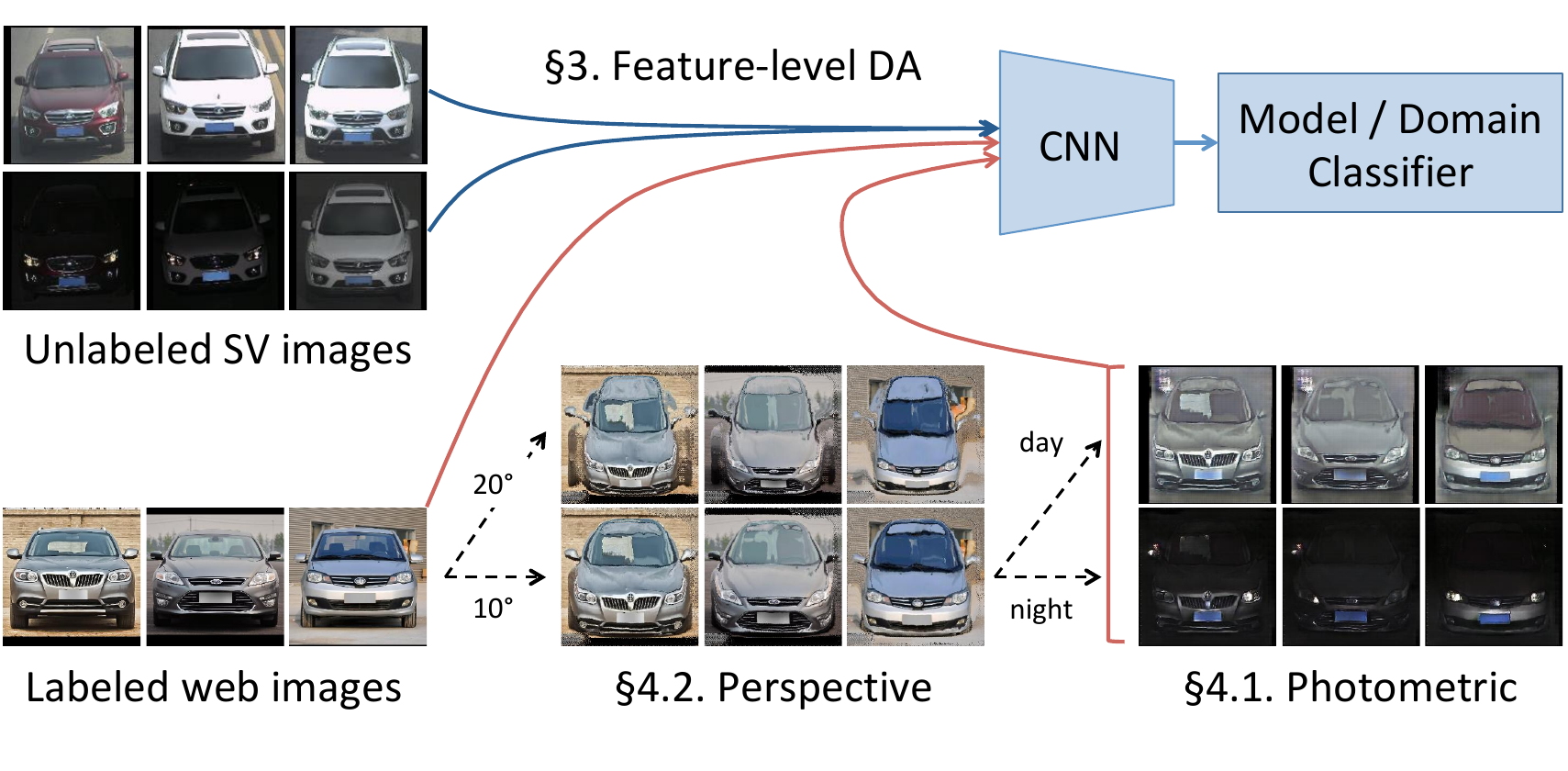}
\vspace{-0.15in}
\caption{Overview of our car recognition system using labeled web and unlabeled surveillance (SV) images. Images taken by SV cameras are different from web images in nameable factors, such as viewpoint or lighting conditions as well as other nuisance factors. We integrate pixel-level DA for perspective and photometric transformations and feature-level DA for other nuisance factors.\label{fig:teasar}}
\vspace{-0.2in}
\end{figure}

A challenge for pixel-level DA is to simultaneously transform source image properties at multiple semantic levels. In Section~\ref{sec:appflow}, we present pixel-level DA by image transformations that make use of vision concepts to deal with different variation factors, such as photometric or geometric transformations (\Fig{fig:teasar}),\footnote{Our framework is unsupervised DA in the sense that we don't require recognition labels from the target domain for training, but it uses side annotations to inject insights from vision concepts for pixel-level adaptation.} for recognition in surveillance domain. To handle low-level transformations, we propose an attribute-conditioned CycleGAN ({\accgan}) that extends \cite{zhu2017unpaired} to generate multiple target images with different attributes. To handle high-level identity-preserving pose transformations, we use an appearance flow (AF)~\cite{zhou2016view}, an warping-based image synthesis tool. To reduce semantic gaps between synthetic and real images, we propose a generalization of AF with 2D keypoints~\cite{li2016deep} as a domain bridge.

In Section~\ref{sec:exp}, we evaluate our framework on car recognition in surveillance images from the comprehensive cars (CompCars) dataset~\cite{yang2015large}. We define an experimental protocol with web images as labeled source domain and surveillance images as unlabeled target domain. We explicitly handle nameable factors of variation such as pose and lighting through pixel-level DA, while other nuisance factors are handled by feature-level DA. As in Table~\ref{fig:teasar_small}, we achieve $\mathbf{84.20}\%$ accuracy, reducing error by $\mathbf{64.9}\%$ from a model trained only on the source domain. We present ablation studies to demonstrate the importance of each adaptation component by extensively evaluating performances with various mixtures of components. We further validate the effectiveness of our proposed feature-level DA methods on standard UDA benchmarks, namely digits and traffic signs~\cite{ganin2016domain} and office-31~\cite{saenko2010adapting}, achieving state-of-the-art recognition performance.

In summary, the contributions of our work are:
\vspace{-0.25cm}
\begin{tight_itemize}
\item{A novel UDA framework that adapts at multiple semantic levels from feature to pixel, with complementary insights for each type of adaptation.}
\item{For feature-level DA, a connection of DANN to a semi-supervised variant, motivating a novel regularization via classification-aware domain adversarial neural network.}
\item{For pixel-level DA, an attribute-conditioned CycleGAN to translate a source image into multiple target images with different attributes, along with an warping-based image synthesization for identity-preserving pose translations via a keypoint-based appearance flow.}
\item{A new experimental protocol on car recognition in surveillance domain, with detailed analysis of various modules and efficacy of our UDA framework.}
\item{State-of-the-art performance on standard UDA benchmarks, such as office-31 and digits, traffic signs adaptation tasks, with our feature-level DA method.}
\end{tight_itemize}
\vspace{-0.07cm}
Due to a large volume of our work, we put additional detail in Section~\ref{sec:supp_gradient}--\ref{supp:cyclegan} of the supplementary material. 

% ------- related -------
\vspace{0.07in}
\Section{Related Work}
\label{sec:related}
\vspace{-0.02in}
\Paragraph{Unsupervised Domain Adaptation.}
Following theoretical developments of domain adaptation~\cite{ben2007analysis,ben2010theory}, a major challenge is to define a proper metric measuring the domain difference. The maximum mean discrepancy~\cite{uda-long2013,tzeng2014deep,fernando2015joint,tzeng2015simultaneous,sun2016deep}, which measures the difference based on kernels, and the domain adversarial neural network~\cite{ganin2016domain,bousmalis2016domain,Bousmalis_2017_CVPR,Sohn_2017_ICCV,sohn_iclr19}, which measures the difference using discriminator, have been successful. Noticing the similarity in problem settings between UDA and SSL, there have been attempts to combine ideas from SSL. For example, entropy minimization~\cite{grandvalet2005semi} has been used in addition to domain adversarial loss~\cite{long2016unsupervised,luo_nips17_label}. Our feature-level DA is built on DANN by resolving issues of discriminator in discovering modes in the feature space. Our formulation also connects tightly to SSL and we explain why entropy minimization is essential for DANN.

\Paragraph{Perspective Transformation.}
Previous works~\cite{yang2015weakly,kulkarni2015deep,tatarchenko2015single} propose encoder-decoder networks to generate output images of target viewpoint. Adversarial learning for perspective transformation~\cite{tran2017disentangled,tran2018representation,Yin_2017_ICCV} has demonstrated good performance on disentangling viewpoint from other appearance factors, but there are still concept (e.g., class label) switches in unpaired settings. Rather than learning the output distribution, \cite{zhou2016view,Park_2017_CVPR} propose an warping-based viewpoint synthesization by estimating a pixel-level flow field. We extend it to improve generalization to real images using synthetic-to-real domain invariant representations such as 2D key points~\cite{li2016deep}.

\Paragraph{Image-to-image Translation.}
With the success of GAN on image generation~\cite{goodfellow2014generative,radford2015unsupervised}, conditional variants of GAN~\cite{mirza2014conditional} have been successfully adopted to image-to-image translation problems in both paired~\cite{isola2017image} and unpaired~\cite{shrivastava2016learning,taigman2016unsupervised,zhu2017unpaired} training settings. Our model extends the work of \cite{zhu2017unpaired} for image translation in unpaired settings using a control variable or visual attribute~\cite{yan2016attribute2image} to generate multiple outputs.

\Paragraph{Multi-level UDA.}
A combination of pixel and feature level adaptation has been attempted in~\cite{hoffman2017cycada}, however, we differ in a few important ways. Specifically, we go further in using insights from SSL that allows novel regularization for feature-level DA, while exploiting 3D geometry and attribute-based conditioning in GANs to simultaneously handle high-level pose and low-level lighting variations. Our experiments include a detailed study of the complementary benefits, as well as the effectiveness of various adaptation modules. While \cite{hoffman2017cycada} consider problems such as semantic segmentation, we study a car recognition problem that highlights the need for adaptation at various levels. We also demonstrate state-of-the-art results on standard UDA benchmarks.

% ------- domadapt -------
\vspace{0.05in}
\Section{Domain Adversarial Feature Learning}
\label{sec:domadapt}
This section describes a classification-aware domain adversarial neural network (\Fig{fig:domadv-ss}) that improves upon a domain adversarial neural network~\cite{ganin2016domain} by joint parameterization of classifier and discriminator.

\Paragraph{Notation.}
Let $\mathcal{X}_{\text{S}},\mathcal{X}_{\text{T}}\,{\subset}\,\mathcal{X}$ be source and target datasets and $\mathcal{Y}\,{=}\,\{1,...,\ncls\}$ be the set for class label. Let $f\,{:}\,\mathcal{X}\,{\rightarrow}\,\mathbb{R}^{\ndim}$ be the feature generator, e.g., CNN, with parameters $\theta_{f}$ that maps input $x\,{\in}\,\mathcal{X}$ into a $\ndim$-dimensional vector.

\SubSection{Recap: Domain Adversarial Neural Network}
\label{sec:domadapt-dann}
Domain adversarial training~\cite{ganin2016domain} aims to adapt classifier learned from the labeled source domain to unlabeled target domain by making feature distributions of the two domains indistinguishable. This is achieved through a domain discriminator $D\,{:}\,\mathbb{R}^{\ndim}\,{\rightarrow}\,(0,1)$ that tells whether features from the two domains are still distinguishable. Then, $f$ is trained to confuse $D$ while classifying the source data correctly:

\vspace{-0.12in}
{{\small
\begin{align}
\max_{\theta_{c}}&\{\mathcal{L}_{\text{C}} =\mathbb{E}_{\mathcal{X}_{\text{S}}} \log C(f, y)\}\label{eq:dann-cls}\\
\max_{\theta_{d}}&\{\mathcal{L}_{\text{D}} =\mathbb{E}_{\mathcal{X}_{\text{S}}} \log (1{-}D(f)) + \mathbb{E}_{\mathcal{X}_{\text{T}}} \log D(f)\} \label{eq:dann-disc}\\
\max_{\theta_{f}}&\{\mathcal{L}_{\text{F}} =\mathcal{L}_{\text{C}} + \lambda \mathbb{E}_{\mathcal{X}_{\text{T}}} \log (1{-}D(f))\} \label{eq:dann}
\eqnvspace
\end{align}
}}
\vspace{-0.12in}

{\noindent}$C\,{:}\,\mathbb{R}^{\ndim}\,{\times}\,\mathcal{Y}\,{\rightarrow}\,(0,1)$ is a class score function that outputs the probability of an input $x$ being a class $y$ among $\ncls$ categories, i.e., $C(f(x),y)\,{=}\,P(y|f(x);\theta_{c})$. $\lambda$ balances classification and domain adversarial losses. The parameters $\{\theta_{c},\theta_{d}\}$ and $\{\theta_{f}\}$ are updated in turn using stochastic gradient descent.

\SubSection{Classification-Aware Adversarial Learning}
\label{sec:domadapt-semisup}
We note that the problem setup of unsupervised domain adaptation is not different from that of semi-supervised learning once we remove the notion of domains. Inspired by the semi-supervised learning formulation of GANs~\cite{salimans2016improved,dai2017good}, we propose a new domain adversarial learning objective that jointly parameterizes classifier and discriminator as follows:

\vspace{-0.12in}
{{\small
\begin{align}
\eqnvspace
\max_{\theta_{c}}& \{\overbar{\mathcal{L}}_{\text{C}}=\mathbb{E}_{\mathcal{X}_{\text{S}}}\log \overbar{C}(y) + \mathbb{E}_{\mathcal{X}_{\text{T}}}\log \overbar{C}(\ncls{+}1)\}\label{eq:dann-semisup}\\
\max_{\theta_{f}}& \{\overbar{\mathcal{L}}_{\text{F}}=\mathbb{E}_{\mathcal{X}_{\text{S}}}\log \overbar{C}(y|\mathcal{Y})+\lambda\mathbb{E}_{\mathcal{X}_{\text{T}}}\log (1{-}\overbar{C}(\ncls{+}1))\} \label{eq:dann-semisup-adv}
\eqnvspace
\end{align}
}}
\vspace{-0.12in}

{\noindent}where we omit $f(x)$ from $\overbar{C}(f(x),y)$ for presentation clarity. The score function $\overbar{C}$ is defined on $\mathbb{R}^{\ndim}\,{\times}\,\{1,...,{\ncls{+}1}\}$ and the conditional score $\overbar{C}(y|\mathcal{Y})$ is written as follows:
\begin{equation}
\overbar{C}(y|\mathcal{Y})\,{=}\,\tfrac{\overbar{C}(y)}{1{-}\overbar{C}(\ncls{+}1)}, \forall y\,{\leq}\,\ncls,\; \overbar{C}(\ncls{+}1|\mathcal{Y})\,{=}\,0 \label{eq:cond_class_score}
\end{equation}
The formulation no more has a discriminator, but classifier has one additional output entry for the target domain. We call our model a classification-aware DANN or {\dannssl} as it allows discriminator to access to classifier directly. While \cite{salimans2016improved} has demonstrated an effectiveness of joint parameterization in semi-supervised GANs, it is not clearly explained why it is better. In the following, we aim to explain the advantage of {\dannssl} in the context of feature-level UDA.

\Paragraph{Discriminator Should Know Classification Boundary.}

{\noindent}Mode collapse is a critical issue in adversarial learning. To prevent it, discriminator needs to discover as many modes in data distribution as possible. While it is difficult to describe the modes in the input space for generative modeling~\cite{goodfellow2014generative}, it is relatively easy to characterize the modes in the feature space: there are $\ncls$ major modes, each of which corresponds to each output class, and the discriminator is demanded for discovering these modes in the feature space. Unfortunately, the discriminator of DANN is trained with binary supervision, implying that the mode discovery is done unsupervisedly. On the other hand, the modes are already embedded in the discriminator of {\dannssl} via joint parameterization and the adversarial learning can be made easier.

We further investigate the gradient of adversarial loss in \eqref{eq:dann} and \eqref{eq:dann-semisup-adv} with respect to $f$. For the ease of presentation, we assume linear classifier and discriminator. Complete derivation including non-linear version is in Section~\ref{sec:supp_gradient}.
\begin{align}
\tfrac{\partial\log(1{-}{D}(f))}{\partial f} &\;{\text{=}} -{D}(f)w_{d}\label{eq:gradient-dann-ss}\\
\tfrac{\partial\log(1{-}\overbar{C}({\ncls{+}1}))}{\partial f} &\;{\text{=}} -\overbar{C}({\ncls{+}1})(w_{{\ncls{+}1}}{-}\textstyle\sum_{y=1}^{\ncls}w_{y}\overbar{C}(y|\mathcal{Y}))\nonumber
\end{align}
where $w_{d},w_{y}\,{\in}\,\mathbb{R}^{\ndim},y\,{\in}\,\{1,...,\ncls{+}1\}$ are discriminator and classifier weights, respectively. As is evident from \eqref{eq:gradient-dann-ss}, the adversarial loss of DANN cannot capture multiple modes as all target examples induce the gradient of the same direction. Even if we use MLP discriminator in practice, it still demands to discover modes correspond to classes without supervision. The joint parameterization allows not only to push features away from the target domain, but also guides them to be pulled close to classes based on the conditional score $\overbar{C}(y|\mathcal{Y})$ of individual target examples.

\Paragraph{Relation to DANN~\cite{ganin2016domain}.}

{\noindent}Besides parameterization, the learning objectives are tightly linked to those of DANN~\cite{ganin2016domain}. For instance, $\overbar{\mathcal{L}}_{\text{F}}\,{=}\,{\mathcal{L}}_{\text{F}}$ with $D\,{=}\,\overbar{C}(\ncls{+}1)$ and $C(y)\,{=}\,\overbar{C}(y|\mathcal{Y})$. It is also easy to show $\overbar{\mathcal{L}}_{\text{C}}\,{=}\,\mathcal{L}_{\text{C}}\,{+}\,\mathcal{L}_{\text{D}}$ by rewriting $\overbar{C}(y)$ using \eqref{eq:cond_class_score} as follows:
\begin{align}
\overbar{\mathcal{L}}_{\text{C}}\,{=}\,&\mathbb{E}_{\mathcal{X}_{\text{S}}}\log \overbar{C}(y|\mathcal{Y})\,{+}\,\nonumber\\
&\mathbb{E}_{\mathcal{X}_{\text{S}}}\log (1{-}\overbar{C}(\ncls{+}1))\,{+}\,\mathbb{E}_{\mathcal{X}_{\text{T}}}\log \overbar{C}(\ncls{+}1)
\end{align}

\Paragraph{Relation to Maximum Classifier Discrepancy~\cite{Saito_2018_CVPR}.}

{\noindent}We also relate our proposed {\dannssl} to recently proposed maximum classifier discrepancy (MCD) learning for UDA. MCD learns shared feature extractor by reducing the prediction discrepancy between two (or more) maximally different classifiers. We show that our {\dannssl} can be understood as MCD with choices of classifiers and the divergence. Following~\cite{Saito_2018_CVPR}, we define the two classification distributions:
\begin{equation}
p_{1}(y|x_{t}) \,{=}\, \overbar{C}(y|\mathcal{Y}),\, p_{2}(y|x_{t}) \,{=}\, \overbar{C}(y),\, y\,{\leq}\,\ncls{+}1
\end{equation}
Note that two classifiers $F_{1}$ and $F_{2}$ in~\cite{Saito_2018_CVPR} are both represented as $(\ncls{+}1)$-way classifier. Using KL divergence, we obtain following discrepancy loss:
\begin{equation}
-\text{KL}(p_{1}\Vert p_{2}) = \log(1{-}\overbar{C}(\ncls{+}1))\label{eq:relation_to_MCD}
\end{equation}
which is equivalent to the adversarial loss in \eqref{eq:dann-semisup-adv}. This analysis provides a unified view of DANN, MCD and more general class of consistency-based SSL algorithms~\cite{laine2016temporal,tarvainen2017mean,french2017self}. A theoretical comparison of UDA algorithms is important as empirical comparison could sometimes be misleading~\cite{oliver2018realistic}. A full derivation of \eqref{eq:relation_to_MCD} and analysis are in Section~\ref{sec:supp_MCD}.

% ------- appflow -------

\vspace{0.03in}
\Section{Pixel-level Cross-Domain Image Translation}
\label{sec:appflow}
As is common for neural networks, DANN is a black-box algorithm and adding domain-specific insight is non-trivial. On the other hand, certain challenges in DA can be better handled in image space. In this section, we introduce complementary tools to deal with nameable factors of variation, such as photometric or perspective transformations, at the pixel level. To achieve this, we propose extensions to prior works on CycleGAN \cite{zhu2017unpaired} and appearance flows \cite{zhou2016view}. We describe with an illustrative application of car recognition in surveillance domain where the only labeled data is from web domain. The pipeline of our system is in \Fig{fig:teasar}.

\SubSection{Photometric Transformation by CycleGAN}
\label{sec:domadapt_pix}
As noticed from \Fig{fig:teasar}, images from surveillance domain have disparate color statistics from web images as they might be acquired outdoors at different times with significant lighting variations. CycleGAN~\cite{zhu2017unpaired} is proposed as a promising tool for image translation by disentangling low-level statistics from geometric structure. A limitation, however, is that it generates a single output when there could be multiple output styles. We propose an attribute-conditioned CycleGAN ({\accgan}) that generates diverse output images with the same geometric structure by incorporating a conditioning variable into generators.

Let $\mathcal{A}$ be a set of attributes in the target domain (day or night). We learn a generator $G\,{:}\,\mathcal{X}_{\text{S}}\,{\times}\,\mathcal{A}\,{\rightarrow}\,\mathcal{X}_{\text{T}}$ that translates an image with certain style $a\,{\in}\,\mathcal{A}$ by fooling an attribute-specific discriminator $D_{a}$. The learning objectives are:

\vspace{-0.12in}
{{\small
\begin{align}
\max_{\theta_{d_{a}}}\{\mathcal{L}_{D_{a}} & {=}\mathbb{E}_{\mathcal{X}_{\text{T}_{a}}}{\log}D_{a}(x) {+} \mathbb{E}_{\mathcal{X}_{\text{S}}}\log (1{-}D_{a}(G(x,a)))\}\\
\max_{\theta_{g}}\{\mathcal{L}_{G} & {=}\mathbb{E}_{\mathcal{X}_{\text{S}}}\mathbb{E}_{\mathcal{A}}{\log}D_{a}(G(x,a))\}
\end{align}
}}
\vspace{-0.12in}

{\noindent}We use multiple discriminators to prevent competition between different attribute configurations, but it is feasible to have one discriminator with $(|\mathcal{A}|{+}1)$-way domain classification loss~\cite{taigman2016unsupervised}. Also, one might afford to have multiple generators per attribute without sharing parameters.\footnote{Empirically, using two separate generators for day and night performs slightly better than a single generator. Please see Section~\ref{supp:cyclegan} for results.} Following~\cite{zhu2017unpaired}, we add cycle consistency loss as follows:
\begin{equation}
\mathbb{E}_{\mathcal{X}_{\text{S}}}\Vert F(G(x{,} a){,}a){-}x\Vert_{1} {+} \mathbb{E}_{\mathcal{X}_{\text{T}_{a}}}\Vert G(F(x{,}a){,}a){-}x\Vert_{1}
\end{equation}
where an inverse generator $F$ maps outputs back to source domain $F(G(x, a), a)\,{=}\,x$. We also use patchGAN~\cite{isola2017image,zhu2017unpaired} for discriminators that makes real or fake decisions from local patches and UNet~\cite{isola2017image} for generators, each of which contributes to preserve geometric structure of an input image.

\SubSection{Perspective Synthesis by Appearance Flow}
\label{sec:domadapt_af}
Besides color statistics, we observe significant differences in camera perspective (\Fig{fig:teasar}). In this section, we deal with perspective transformation using an image warping based on a pixel-wise dense flow called appearance flow (AF)~\cite{zhou2016view}. Specifically, we propose to improve the generalization of AF estimation network (AFNet) trained on 3D CAD rendered images to real images by utilizing a robust representation across synthetic and real domains, i.e. 2D keypoints.

\Paragraph{Appearance Flow.}

{\noindent}Zhou et al.~\cite{zhou2016view} propose to estimate a pixel-level dense flow from an input image with target viewpoint and synthesize an output by reorganizing pixels using bilinear sampling~\cite{jaderberg2015spatial}:
\begin{equation}
I_{p}^{i,j}{=}\textstyle\sum_{(h,w) \in N} I_s^{h,w}(1{-}|F_y^{i,j}{-}h|) (1{-}|F_x^{i,j}{-}w|),\label{eqn:af}
\end{equation}
where $I_{s},I_{p}$ are input and output, $(F_{x}, F_{y})$ is a pixel-level flow field in horizontal and vertical axes called appearance flow (AF), estimated by an AF estimation network (AFNet). $N$ denotes 4-pixel neighborhood of ($F_{x}^{i,j}, F_{y}^{i,j}$). In contrast to neural network based image synthesization methods~\cite{tatarchenko2015single}, AF-based transformation may have a better chance of preserving object identity since all pixels of an output image are from an input image and no new information, such as learned priors in the decoder network, is introduced.

\begin{figure}[t!]
\centering
\includegraphics[width=0.85\linewidth]{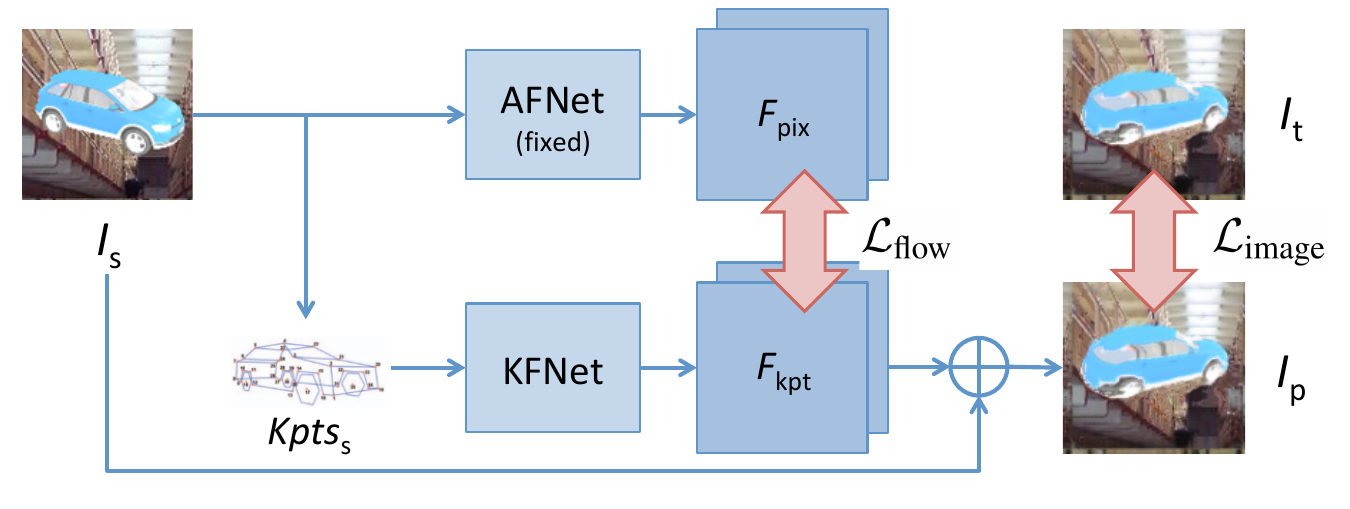}
\vspace{-0.1in}
\caption{Training framework of keypoint-based appearance flow network (KFNet) by distilling knowledge from pretrained AFNet.}
\label{fig:kfnet}
\vspace{-0.2in}
\end{figure}

\Paragraph{Keypoint-based Robust Estimation of AF.}
\label{sec:appflow-kpt}

{\noindent}AFNet requires image pairs $(I_{s}, I_{t})$ with perspective being the only factor of variation for training. Since it is infeasible to collect precisely controlled dataset of real images at large-scale, rendered images from 3D CAD models are used. However, this induces a generalization issue when applied to real images at test time.

To make AFNet generalizable, we propose sparse 2D keypoints in replace of an RGB image as an input to AFNet both at train and test times. Although sparse, for objects like cars, we argue that 2D keypoints contain sufficient information to reconstruct (rough) geometry of an entire object, while being invariant across rendered and real domains. Besides, keypoint estimation can be done robustly across synthetic and real domains even when the keypoint localization network is trained only on the synthetic data~\cite{li2016deep}. To this end, we propose a 2D keypoint-based AFNet (KFNet) that takes estimated 2D keypoints and the target viewpoint as an input pair to generate flow fields $F$ for synthesization.

The KFNet is trained using rendered image pairs. Moreover, we leverage pretrained AFNet that produces a robust AF representation for rendered images to train the KFNet by distillation. The learning objective is as follows:
\begin{equation}
\min\{\mathcal{L} = \Vert F_{\text{kpt}} - F_{\text{pix}}\Vert_{1} + \lambda\Vert I_{p}(F_{\text{kpt}}, I_{s}) - I_{t} \Vert_{1}\}\label{eqn:kf}
\end{equation}
where $F_{\text{kpt}}$ is an estimated appearance flow by KFNet and $F_{\text{pix}}$ is that by AFNet. Here, $I_{p}(F, I_{s})$ is the predicted image from $I_{s}$ using $F$ based on \eqref{eqn:af}. The training framework by distillation is visualized in \Fig{fig:kfnet}.

% ------- experiment -------
\vspace{0.05in}
\Section{Experiments}
\label{sec:exp}
We strive for providing empirical evidence for the effectiveness of individual components of our proposed framework as well as their complementarity by conducting extensive experiments on car recognition in surveillance domain. For feature-level adaptation, we also provide performance comparison on standard benchmarks, namely digits and traffic signs~\cite{ganin2016domain} and office-31~\cite{saenko2010adapting}.

\SubSection{Car Recognition in Surveillance Domain}
\label{sec:exp-setting}
\vspace{-0.05in}
\Paragraph{Dataset.}
CompCars dataset~\cite{yang2015large} offers two datasets, one from the web and the other from the surveillance (SV) domains. It contains $52,083$ web images across $431$ car models and $44,481$ SV images across $181$ car models. Samples are in \Fig{fig:teasar}. The SV test set contains $9,630$ images across $181$ car models, of which $6,404$ images are in day condition.\footnote{We provide a binary label (day or night) for images from surveillance domain by computing the mean pixel-intensity.}

To train an appearance flow estimation network, based on emperical distribution of web images, we render car images at multiple elevation ($0^{\circ}{\sim}30^{\circ}$) and azimuth variations ($\pm 15^{\circ}$) from ShapeNet~\cite{shapenet2015}. We apply pixel-level adaptation to $5,508$ web images of frontal view. 

\Paragraph{Training.}
The task is to train a classifier that works well on SV images using labeled web (source) and unlabeled SV (target) images. We use ResNet-18~\cite{he2016deep} fine-tuned on web images as our baseline. Then, we train models with different integration of pixel and feature-level DA components. Note that synthesized images by pixel-level adaptation are considered as labeled training examples. Furthermore, we use data augmentation, such as translation, horizontal flip or chromatic jitter, for all models by default. We refer to Section~\ref{sec:supp_implementation_details_dann} for more training details.

\Paragraph{Model Selection.}
While it is desirable to do a model selection without labeled examples from the target domain, to our knowledge, there does not exist an unsupervised evaluation measure that is highly correlated with the supervised performance~\cite{bousmalis2016domain}. To allow more meaningful and interpretable comparisons across different methods, we report our results based on a supervised model selection~\cite{bousmalis2016domain} using a small validation set containing approximately $5$ labeled examples per class from the target domain. We provide a comprehensive comparison to unsupervised model selection using a variant of reverse validation~\cite{zhong2010cross,ganin2016domain} in Section~\ref{sec:supp-model-selection}.

\begin{table}[t]
\footnotesize
\centering
\begin{tabular}{ccccc}
\toprule
ID & Perspective Transformation & SV & Day & Night \\
\midrule
M1 & Baseline (web only) & 54.98 & 72.67 & 19.87 \\
M3 & Appearance Flow (AF) & 59.73 & 75.78 & 27.87 \\
M4 & Keypoint-based AF (KF) & 61.55 & 77.98 & 28.92 \\
M5 & KF with mask (MKF) & 64.30 & 78.62 & 35.87 \\
\bottomrule
\end{tabular}
\vspace{-0.08in}
\caption{Accuracy on SV test set with different perspective transformation methods: appearance flow (AF), keypoint-based AF (KF) and with mask (MKF).\label{tab:persp_recog}}
\vspace{-0.02in}
\end{table}

\begin{table}[t]
\centering
\footnotesize
\begin{tabular}{ccccc}
\toprule
ID & Photometric Transformation & SV & Day & Night \\
\midrule
M1 & {Baseline (web only)} & 54.98 & 72.67 & 19.87 \\
M6 & {CycleGAN} & 64.32 & 77.01 & 39.12 \\
M7 & {\accgan}  & 67.30 & 78.20 & 45.66 \\
M8 & MKF+CycleGAN & 71.21 & 81.54 & 50.68 \\ 
M9 & MKF+{\accgan} & 79.71 & 84.10 & 70.99 \\ 
\bottomrule
\end{tabular}
\vspace{-0.08in}
\caption{Accuracy on SV test set with different photometric transformation methods: CycleGAN~\cite{zhu2017unpaired}, attribute-conditioned CycleGAN ({\accgan}), and combinations with MKF.\label{tab:photo_recog}}
\vspace{-0.02in}
\end{table}

\begin{table}[t]
\footnotesize
\centering
\begin{tabular}{cccccc}
\toprule
ID & Pixel & Feature & SV & Day & Night \\ \midrule
M1 & \multicolumn{2}{c}{Baseline (web only)} & 54.98 & 72.67 & 19.87 \\ 
M2 & \multicolumn{2}{c}{Supervised (web+SV)} & 98.63 & 98.92 & 98.05 \\ 
M10 &  -- & DANN & 60.40 & 75.56 & 30.31 \\
\cite{hoffman2017cycada} &  CycleGAN & DANN & 64.82 & 76.35 & 41.93 \\
M11 &  -- & {\dannssl} & 75.83 & 76.73 & 74.05 \\
M12 & MKF & {\dannssl} & 80.40 & 82.50 & 76.22 \\
M13 & {\accgan} & {\dannssl} & 80.24 & 82.15 & 76.44 \\
M14 & MKF+{\accgan} & {\dannssl} & \textbf{84.20} & \textbf{85.77} & \textbf{81.10} \\
\bottomrule
\end{tabular}
\vspace{-0.08in}
\caption{Accuracy on SV test set with pixel and feature-level DA components. We consider an MKF for perspective and attribute-conditioned CycleGAN (AC-CGAN) for photometric transformations for pixel-level DA, and {\dannssl} for feature-level DA.\label{tab:recog}}
\vspace{-0.1in}
\end{table}

\begin{figure}[t]
\centering
\includegraphics[width=0.95\linewidth]{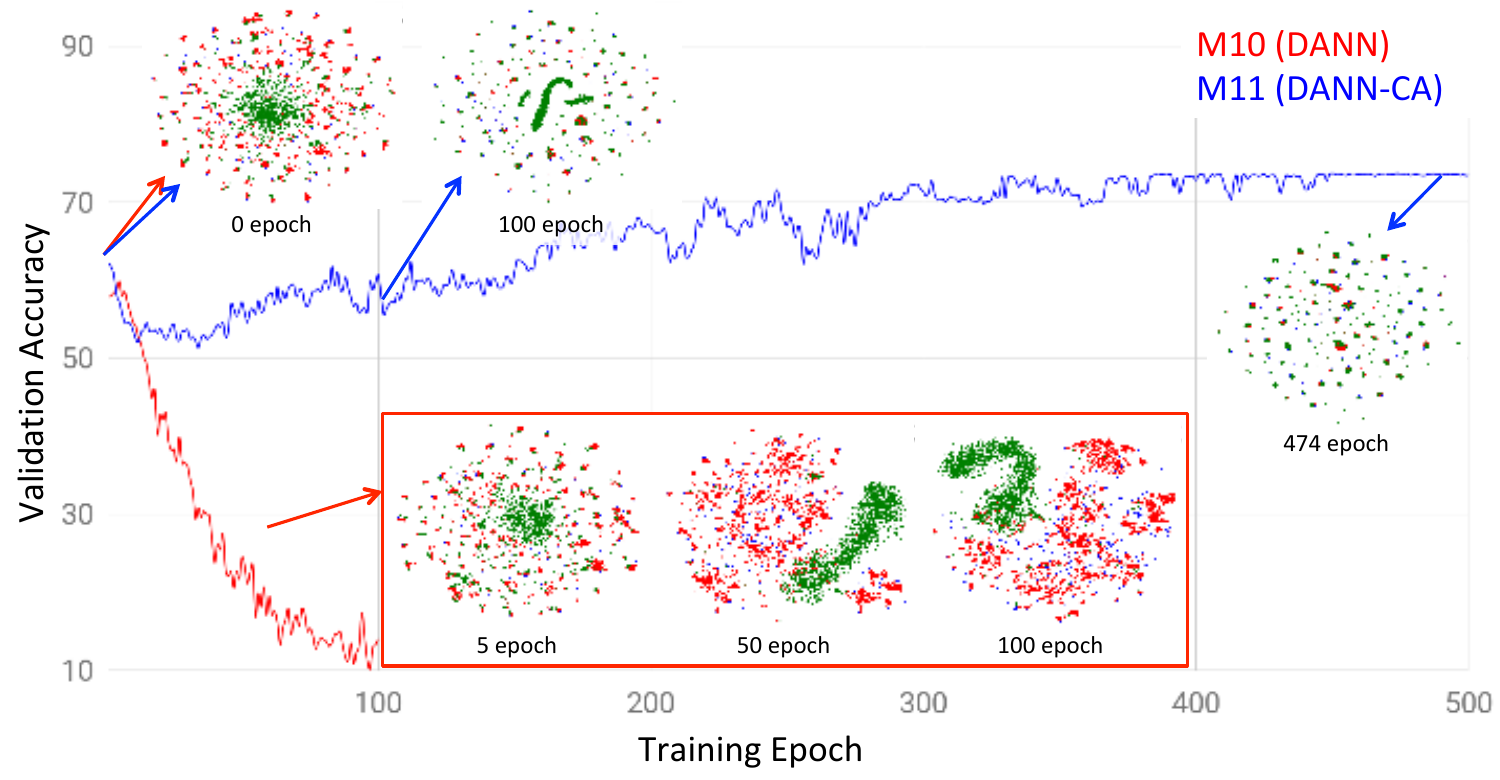}
\vspace{-0.1in}
\caption{Accuracy of DANN ({\color{red}M10}) and {\dannssl} ({\color{blue}M11}) on SV validation set over training. We also visualize t-SNE plots of each model at different training epochs. \label{fig:dann-vs-dann-ca}}
\vspace{-0.05in}
\end{figure}

\SubSection{Summary Results}
We report the classification accuracy on the surveillance test set in \Cref{tab:persp_recog,tab:photo_recog,tab:recog}. Noticing a huge accuracy drop on night images, we also report accuracy of individual day and night sets. We present t-SNE~\cite{van2014accelerating} plots of web (blue), day (red) and night (green) images in \Fig{fig:dann-vs-dann-ca} and \Fig{fig:stability}.

Firstly, although achieving state-of-the-art accuracy on the web test set ($96.4\%$ vs $91.2\%$~\cite{yang2015large}), the baseline model trained only on web images suffers from generalization to SV images, resulting in only $54.98\%$ accuracy. Comparing to the performance of the model trained with target domain supervision ($98.65\%$ in Table~\ref{tab:recog}) provides a sense of how different two domains are. While the baseline adaptation model, DANN (M10 in Table~\ref{tab:recog}), achieves only $58.80\%$, the proposed joint pixel and feature-level adaptation method achieves $\mathbf{84.20}\%$, reducing the error by $\mathbf{64.9}\%$ from the baseline M1. While the use of baseline pixel (CycleGAN) and feature-level (DANN) DA methods as in~\cite{hoffman2017cycada} demonstrates moderate improvement ($64.82\%$) over the baseline, this is far below our proposed DA framework. In the following, we present comprehensive studies on the contribution of individual components and their complementarity.

\begin{table}[t]
\centering
\scriptsize
\begin{tabular}{lcccccc}
\toprule
Method & M$\rightarrow$MM & S$\rightarrow$S & S$\rightarrow$M & M$\rightarrow$S & S$\rightarrow$G \\ \midrule
Source only & 67.90 & 87.05 & 63.74 & 62.44 & 94.53 \\
DANN & \textbf{98.00} & 92.24 & 88.70 & 82.30 & 97.38 \\
{\dannssl} & \textbf{98.03} & \textbf{94.47} & \textbf{96.23} & \textbf{87.48} & \textbf{98.70} \\
\bottomrule
\end{tabular}
\vspace{-0.08in}
\caption{Evaluation on UDA tasks~\cite{ganin2016domain}, such as MNIST to MNIST-M (M$\rightarrow$MM), Synthetic Digits to SVHN (S$\rightarrow$S), SVHN to MNIST (S$\rightarrow$M), MNIST to SVHN (M$\rightarrow$S), or Synthetic Signs to GTSRB (S$\rightarrow$G). Test set accuracy averaged over 10 runs is reported. The best performers and the ones within standard error are bold-faced. \label{tab:dann-comparison-digits}} 
\vspace{-0.02in}
\end{table}

\begin{table}[t]
\centering
\scriptsize
\begin{tabular}{lcccccc}
\toprule
Method & A$\rightarrow$W & D$\rightarrow$W & W$\rightarrow$D & A$\rightarrow$D & D$\rightarrow$A & W$\rightarrow$A \\ \midrule
Source only & 76.42 & 96.76 & 97.99 & 79.81 & 60.44 & 59.53 \\
DANN & 85.97 & 96.87 & 97.94 & 84.12 & 67.63 & 66.78 \\
{\dannssl} & \textbf{91.35} & \textbf{98.24} & \textbf{99.48} & \textbf{89.94} & \textbf{69.63} & \textbf{68.76} \\
\bottomrule
\end{tabular}
\vspace{-0.08in}
\caption{Evaluation on office-31 benchmark~\cite{saenko2010adapting} between Amazon (A), DSLR (D), and Webcam (W) domains using ResNet-50. Target domain accuracy averaged over 5 runs is reported. The best performers and the ones within standard error are bold-faced. \label{tab:dann-comparison-office}} 
\vspace{-0.1in}
\end{table}

\SubSection{Analysis on Pixel-level Adaptation}
\label{sec:exp-pixel}
This section contributes to the analysis of our pixel-level DA on dealing with perspective and photometric transformations, typical factors of variation introduced in SV domain.

\begin{figure*}[t]
\begin{center}
\subfigure[Persp. on rendered data]{\includegraphics[height=1.3in]{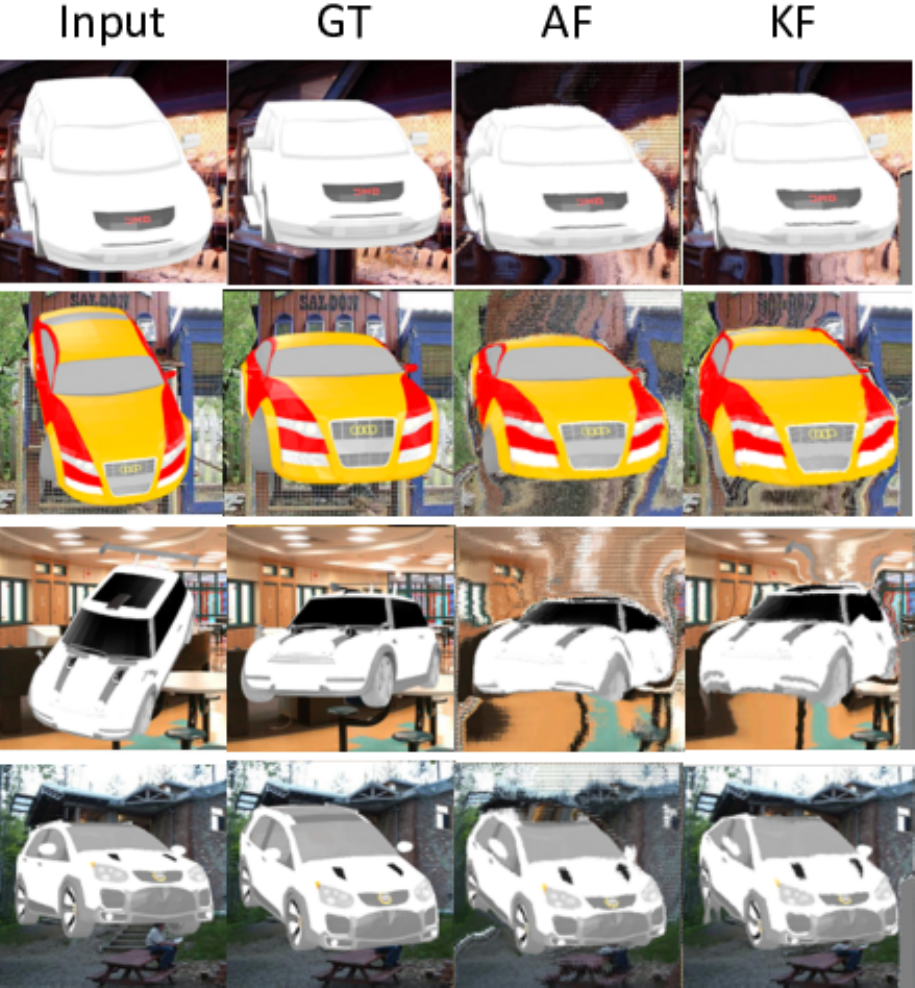}\label{fig:af_syn_visual}}\hspace{0.06in}
\subfigure[Perspective ($0^\circ{\sim}30^\circ$) and photometric (day, night) transformations on real data]{\includegraphics[height=1.3in]{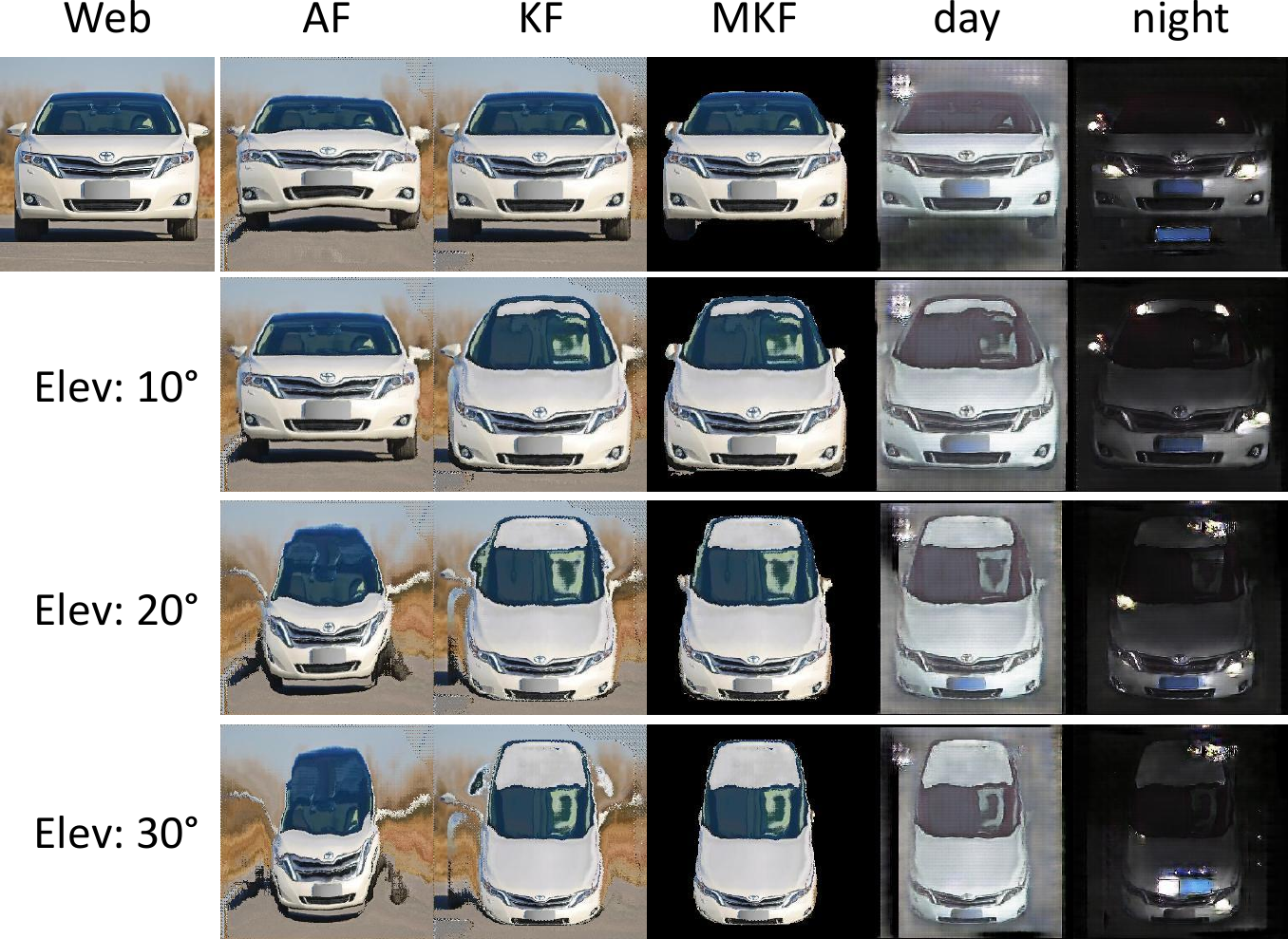}\hspace{0.0in}
\includegraphics[height=1.3in]{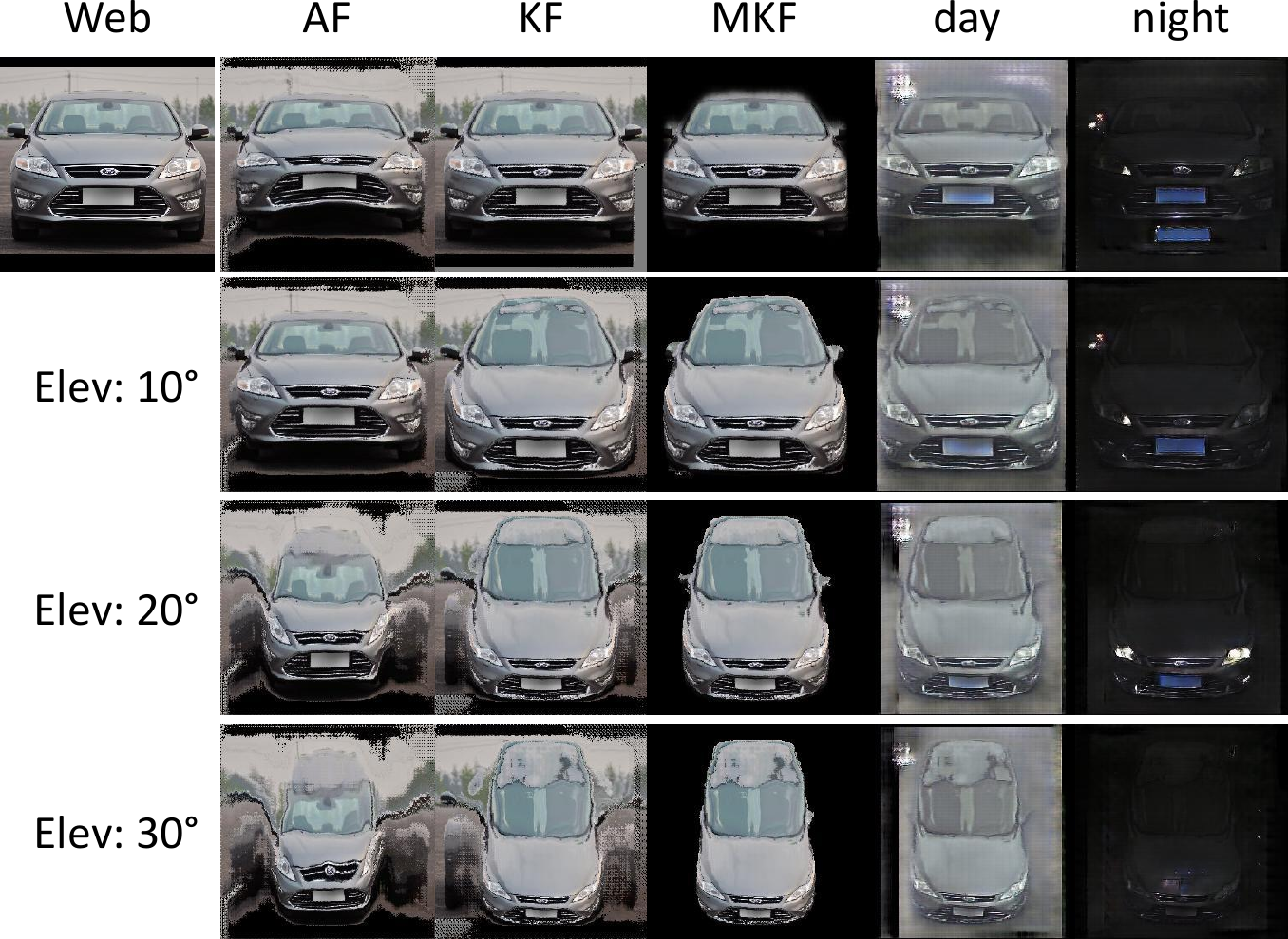}\hspace{0.0in}
\includegraphics[height=1.3in]{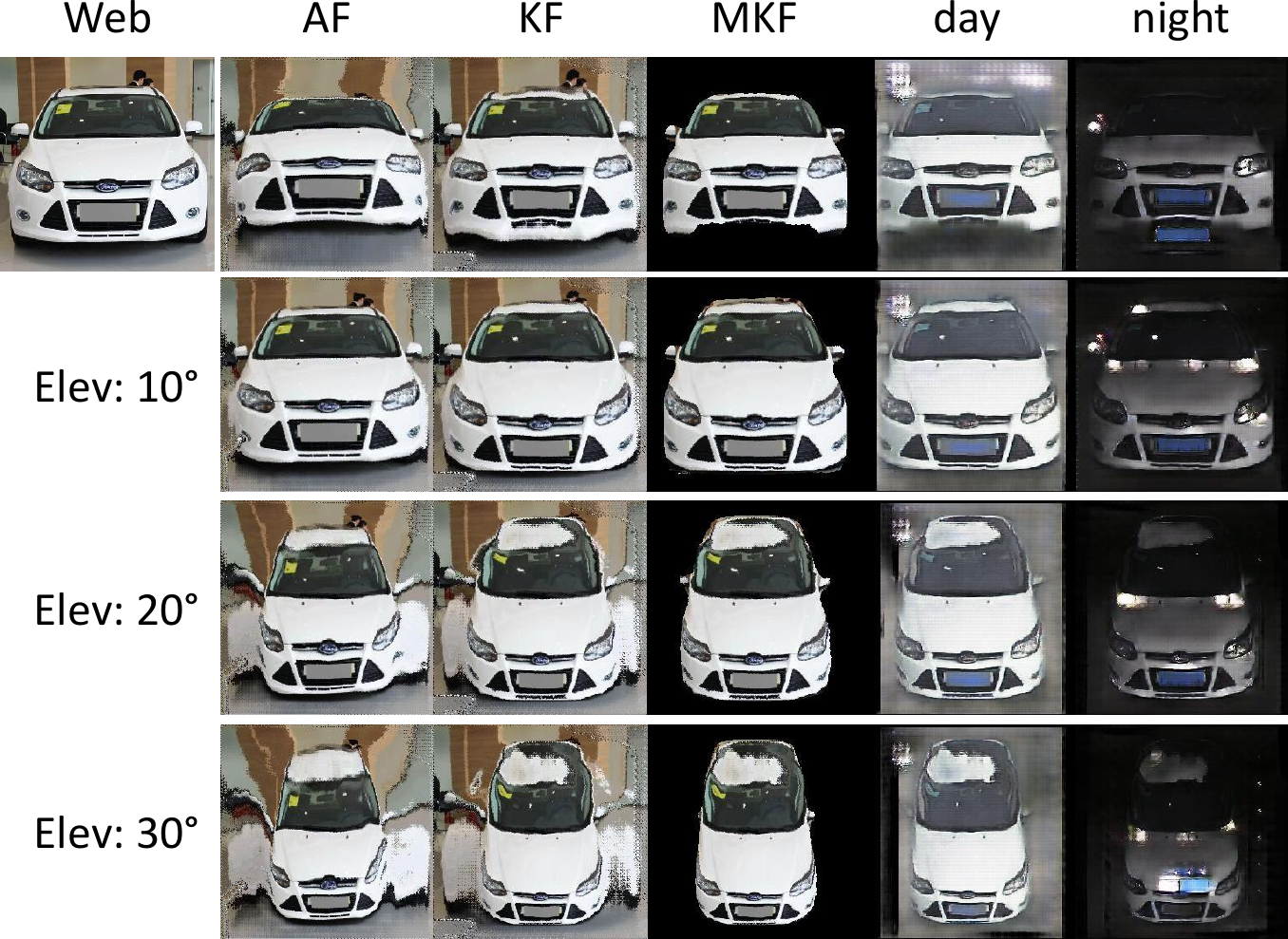}\label{fig:af_visual}}
\vspace{-0.15in}
\caption{Synthesized images by (a) perspective on rendered images of 3D CAD models and (b) perspective and photometirc transformations on real images from CompCars dataset. (a) From left to right: input, GT of target view, and perspective transformed images using AFNet and sparse 2D keypoint-based AFNet (KFNet). (b) From left to right: for each web image, perspective transformed images using AFNet, KFNet and its masked output (MKF), followed by photometric transformation into day and night by {\accgan}.}
\vspace{-0.2in}
\end{center}
\end{figure*}

\Paragraph{Perspective Transformation with CycleGAN~\cite{zhu2017unpaired}.}

\begin{figure}[t]
\centering
\includegraphics[width=0.98\linewidth]{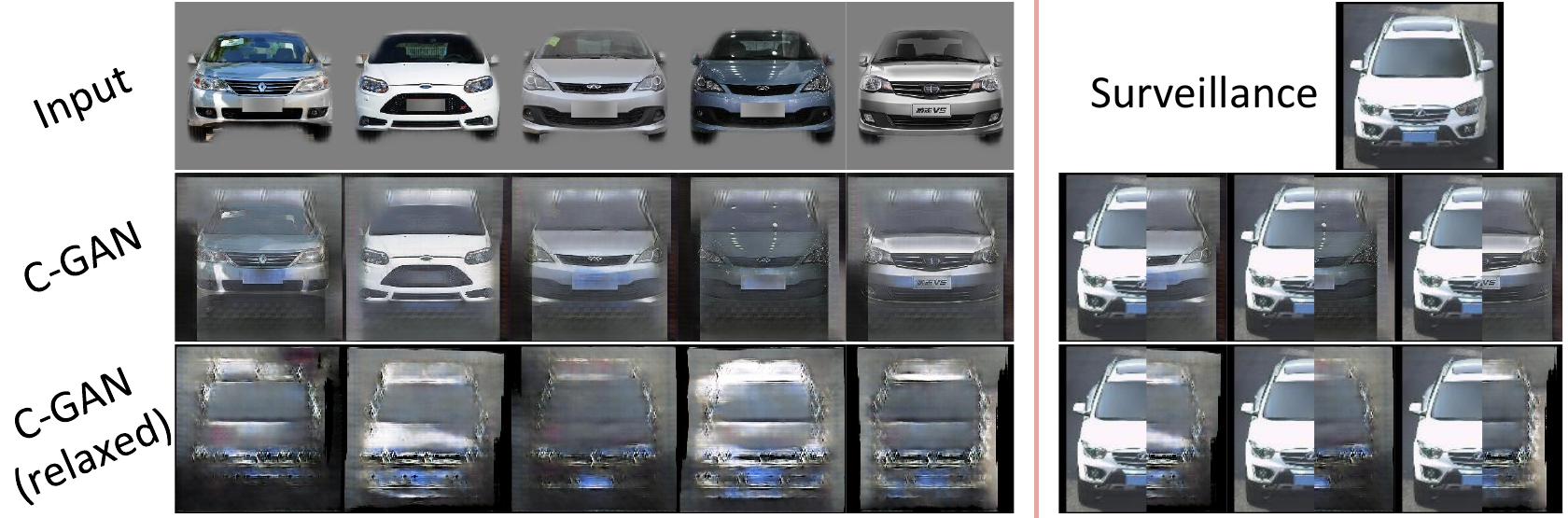}\label{fig:naive_cyclegan_visual}
\vspace{-0.05in}
\caption{Web to SV (day) translation using CycleGAN (second) and its variant (third) by removing skip connection from generator and increasing receptive filed size for patch discriminator. On the right, we overlay left half of translated images with SV image to highlight the impact of constraints on perspective transformation. \label{fig:CycleGAN}}
\vspace{-0.15in}
\end{figure}

{\noindent}The success of CycleGAN on image translation is attributed by few factors, such as cycle consistency loss, patch-based discriminator, or generator with skip connection. However, these constraints may be too strong to translate viewpoint. As is evident from \Fig{fig:CycleGAN}, the output of CycleGAN (second row) maintains the geometric structure of the input (first row) faithfully but fails at adapting to the viewpoint of SV domain. Relaxing constraints, such as removing skip connections of generator and increasing receptive field size of patch-based discriminator, allows perspective adaptation possible (third row), but we lose many details crucial for recognition tasks.

Our approach solves the challenge by translating images in two steps, resulting in high-quality image synthesis from web to SV domain as in \Fig{fig:af_visual}. The conclusion from our visual investigation aligns with the recognition performance, where combined perspective transformation and CycleGAN (M8) achieves $71.21\%$, which improves upon a model without perspective transformation (M6, $64.32\%$) in Table~\ref{tab:photo_recog} or a model without CycleGAN (M5, $64.30\%$) in Table~\ref{tab:persp_recog}. 

\Paragraph{Disentangling Illumination via {\accgan}.}

\begin{figure}[t!]
\begin{center}
\includegraphics[width=0.9\linewidth]{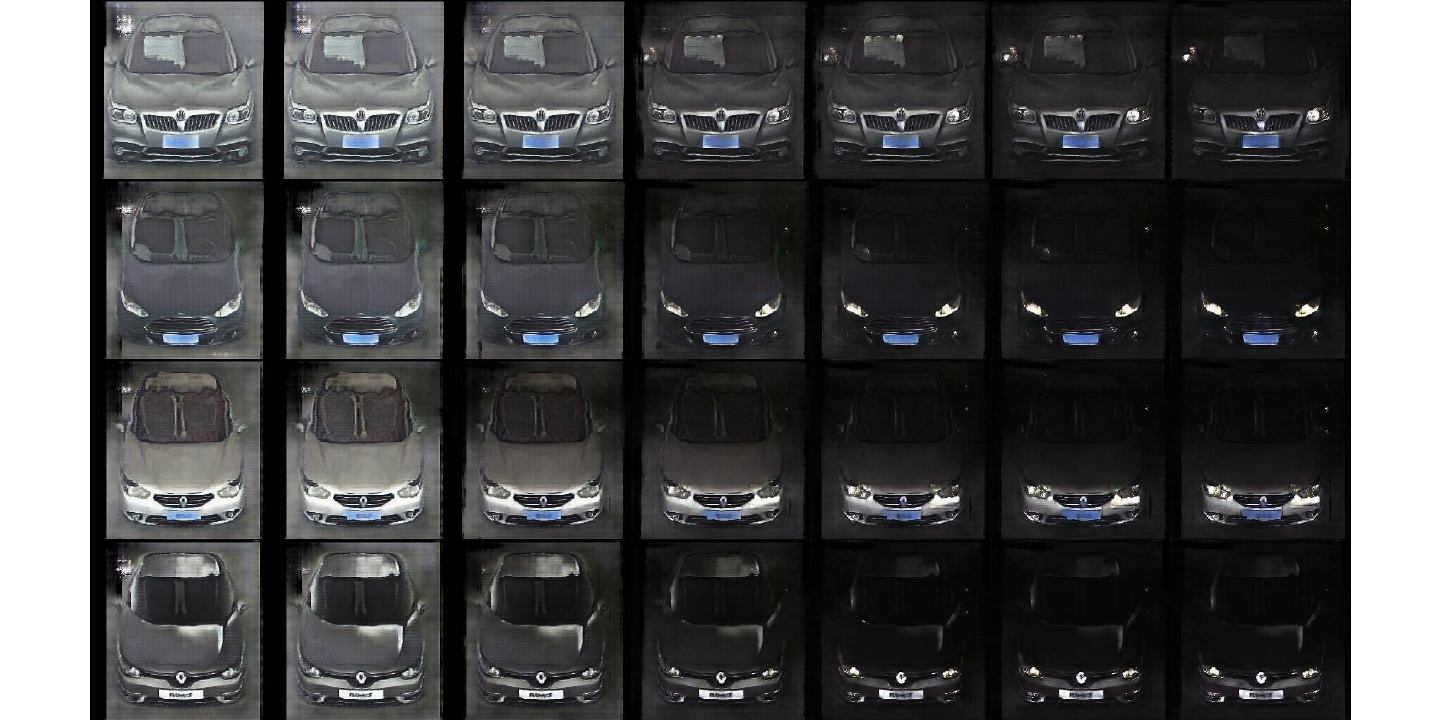}
\vspace{-0.07in}
\caption{Continuous interpolation of latent code of {\accgan}.\label{fig:attribute_interpolation}}
\vspace{-0.28in}
\end{center}
\end{figure}

{\noindent}The {\accgan} fixes the unimodal translation nature of CycleGAN with a latent code~\cite{yan2016attribute2image}. This allows learning disentangled representation from an attribute, which in our case the illumination, and as a result, we can synthesize images of the same car with different illumination conditions, as in \Fig{fig:af_visual}. Moreover, the continuous interpolation of latent code allows to generate continuous change in illumination factor (e.g., color tone, pixel intensity of headlight) without changing the shape and appearance of each car, as in \Fig{fig:attribute_interpolation}.

Generating images with diverse illumination conditions improves the recognition accuracy as in Table~\ref{tab:photo_recog}, especially on the night images of SV domain. The {\accgan} (M7) improves by $2.98\%$ upon the CycleGAN (M6). Moreover, when combined with perspective transformation (M8 and M9), we observe a larger increase in improvement of $8.50\%$.  

\Paragraph{Comparison between AFNet and KFNet.}

{\noindent}KFNet is developed to improve the generalization of AFNet to real images. Before comparing these models on them, we evaluate KFNet on rendered images from 3D CAD models to demonstrate comparable performance to AFNet. We show inputs, output targets and transformed images by AFNet and KFNet in \Fig{fig:af_syn_visual}. We observe reliable estimation of appearance flow by KFNet. Furthermore, we obtain $0.072$ per-pixel L1 reconstruction error between rendered output images and perspective transformed images at four elevations ($0^\circ$ to $30^\circ$) using KFNet, which is comparable to $0.071$ error of AFNet (pixel values are normalized to $[0,1]$).

Now, we show results on real images in \Fig{fig:af_visual}. AFNet struggles to generalize on real images and generates distorted images with incorrect target elevation. Although sparse, 2D keypoints are more robust to domain shift from synthetic to real and are sufficient to preserve the object geometry and correctly transform to the target perspective. Finally, better recognition performance on SV domain of the network trained with source and the perspective transformed images ($59.73\%{\rightarrow}61.55\%$ from M3 to M4 in Table~\ref{tab:persp_recog}) implies the superiority of the proposed KFNet.

\SubSection{Analysis on Feature-level Adaptation}
\label{sec:exp-feat}
We demonstrate the superiority of the proposed {\dannssl} to the DANN on car recognition and other UDA tasks.

\Paragraph{Evaluation on Car Recognition in SV Domain.}

{\noindent}Note that, on top of $512$-dim features, the linear classifier (512$-$431/432) is used for both models, while we use the 3-layer MLP (512$-$320$-$320$-$1) for the discriminator of DANN after trying several discriminator architectures with different depths. As in Table~\ref{tab:recog}, the improvement of {\dannssl} is larger than that of DANN, confirming the superiority of the proposed method. We further investigate the behavior of these methods from training curves in \Fig{fig:dann-vs-dann-ca}. The DANN starts to drop significantly after few epochs of adversarial training, remaining with a few collapsed modes in the end. While it shows some fluctuations at the beginning of training, {\dannssl} shows clear progression over training and finally reaches at convergence.

\Paragraph{Evaluation on UDA Benchmarks.}

{\noindent}We also evaluate the performance of DANN and our {\dannssl} on UDA benchmarks. For digits and traffic signs tasks, we use data augmentation as in~\cite{haeusser2017associative}. Due to space constraint, we provide more details on experimental setting and comparison to other methods in Section~\ref{supp:benchmark_details}. As we see in the summary results of Table~\ref{tab:dann-comparison-digits} and \ref{tab:dann-comparison-office}, our proposed {\dannssl} outperforms the DANN on all tasks and sometimes by a huge margin. We remind that the only difference between the two methods is the parameterization of the classifier and discriminator, and it clearly shows the importance of joint parameterization in adversarial domain adaptation.

\SubSection{Analysis on Joint PnF Adaptation}
\label{sec:exp-joint}
Finally, we provide an empirical analysis on the proposed joint pixel and feature-level (PnF) adaptation. In the joint framework, we train models with feature-level adaptation methods using unlabeled target domain and expanded labeled source domain including original source images and synthesized images by pixel-level DA.

\Paragraph{Improved Domain Alignment with Feature-level DA.}

{\noindent}While it allows high-fidelity generation, constraints in the pixel-level DA make it hard to faithfully adapt to the target domain. It is evident from \Fig{fig:stability} where t-SNE plot of M9 is less clean than that of M11. This implies that the role of feature-level DA in joint DA framework is to learn remaining factors not yet discovered by the pixel-level DA.

\Paragraph{Improved Training Stability with Pixel-level DA.}

{\noindent}We delve deeper into understanding the interplay between pixel and feature-level DAs. \Fig{fig:stability} shows accuracy curves of pixel-level (M9), feature-level (M11) and joint (M14) DA models on day (dotted) and night (solid) of SV validation sets. While the accuracy on days are stable for all models, we observe a large up-and-down for curve on nights of M11. Note that the fluctuation in the night curve of M9 is not as significant. This is due to many constraints (e.g., warp-based viewpoint synthesis, cycle-consistency or UNet architecture) imposed on the training of pixel-level DA, allowing high-fidelity translation of perspective and illumination variations whose outputs are closer to the target domain than the source examples. Consequently, M14 shows significantly less fluctuation during the training than M11.

We further study the training stability from the mode coverage perspective. Assuming modes correspond to classes in the feature space, the number of classes that are not assigned as top-1 prediction by any of SV test set images is used as a proxy to mode coverage. We provide results in Table~\ref{tab:mode_coverage}. While M11 has $29.6$ classes on average over 5 runs with no assigned SV image, only 2 classes are missing for M9. The pixel-level DA effectively complements the mode collapse of adversarial learning in the feature-level DA, reducing the number of missing modes to $10.4$ for M14.

\Paragraph{Complementarity of Components.}

{\noindent}To summarize, each module has its own disadvantage, such as training instability for feature-level DA and the lack of adaptation flexibility for pixel-level DA. Our empirical analysis suggests that these shortages can be complemented when combined in a unified framework, improving the accuracy by $4.49\%$ and $8.37\%$ upon individual modules, respectively.

\definecolor{olivegreen}{HTML}{38761D}

\begin{figure}[t]
\begin{center}
\includegraphics[width=0.85\linewidth]{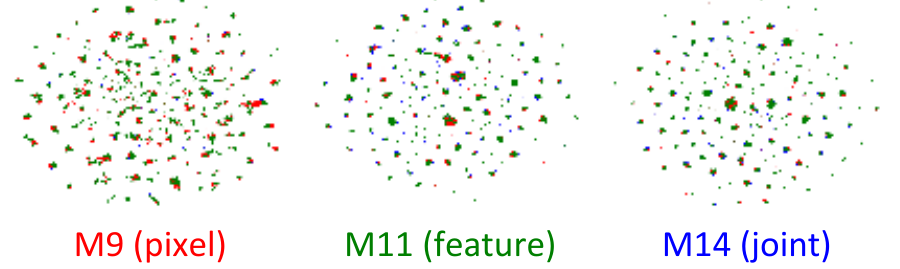}
\includegraphics[width=0.95\linewidth]{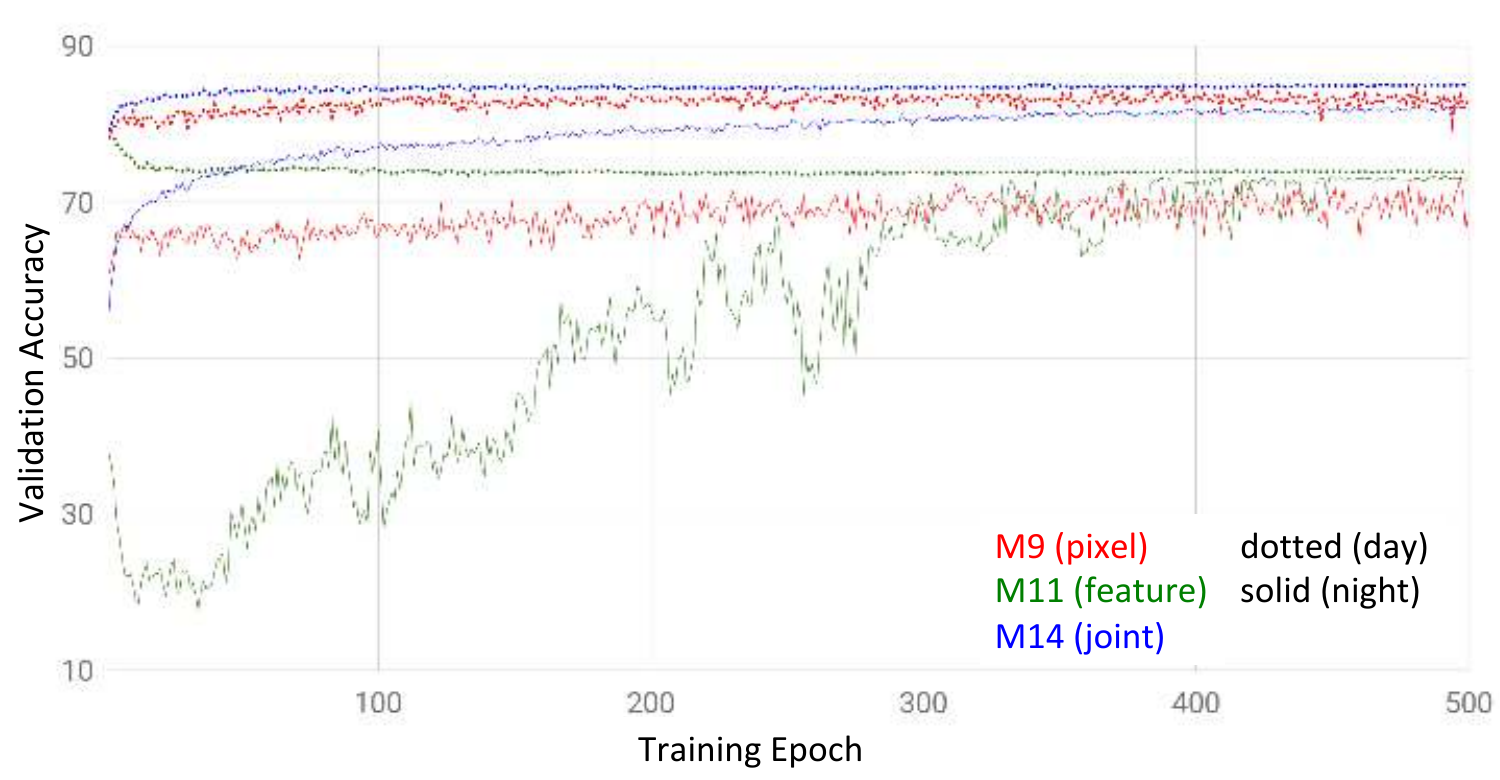}
\vspace{-0.1in}
\caption{Accuracy curves on day (dotted) and night (solid) SV validation set over training and t-SNE plots of pixel-level ({\color{red}M9}), feature-level ({\color{olivegreen}M11}) and joint ({\color{blue}M14}) DA models.}
\label{fig:stability}
\end{center}
\vspace{-0.12in}
\end{figure}

\begin{table}[t]
\footnotesize
\begin{center}
\begin{tabular}{cccc}
\toprule
 & M9 (pixel) & M11 (feature) & M14 (joint) \\
\midrule
\# missing modes & 2 & 29.6$\pm$1.1 & 10.4$\pm$0.6 \\
\bottomrule
\end{tabular}
\end{center}
\vspace{-0.2in}
\caption{Number of missing modes (classes) out of 181 classes.\label{tab:mode_coverage}}
\vspace{-0.2in}
\end{table}

% ------- conclusion -------
\vspace{-0.02in}
\section{Conclusion}
\label{sec:conc}
\vspace{-0.05in}
With an observation that certain adaptation challenges are better handled in feature space and others in pixel space, we propose a joint UDA framework by leveraging complementary tools that are better-suited for each type of adaptation challenge. Importance and complementarity of each component are demonstrated through extensive experiments on a novel application of car recognition in surveillance domain. We also demonstrate state-of-the-art performance on UDA benchmarks with our proposed feature-level DA methods. 

\vspace{-0.06in}
\subsubsection*{Acknowledgement}
\vspace{-0.05in}
\noindent We thank Paul Vernaza, Gaurav Sharma, Wongun Choi, and Wenling Shang for helpful discussions.

{\footnotesize
\bibliographystyle{ieee}
\bibliography{domadapt}
}

% ------- supplementary material -------
\newpage
\appendix
\onecolumn

\renewcommand{\thesection}{S\arabic{section}}   
\renewcommand{\thetable}{S\arabic{table}}   
\renewcommand{\thefigure}{S\arabic{figure}}
\renewcommand{\theequation}{S\arabic{equation}}

% ------- gradient -------
\section{A Gradient Analysis of Classification-Aware DANN}
\label{sec:supp_gradient}
Here we provide a detailed derivation of the gradient analysis of DANN and {\dannssl} presented in Section~\ref{sec:domadapt-semisup}. 

\subsection{Gradient for DANN}
Let $\phi_{d}\,{:}\,\mathbb{R}^{\ndim}\,{\rightarrow}\,\mathbb{R}$ be a function that generates the exponent of discriminator distribution, i.e., $D(f) \,{=}\, \sigma\left(\phi_{d}(f)\right)$ where $\sigma(\cdot)$ is a logistic (sigmoid) function. Then, we get the following gradient:
\begin{align}
\frac{\partial\log(1{-}D(f))}{\partial f} &= \frac{-1}{1{-}D(f)}D(f)(1{-}D(f))\frac{\partial \phi_{d}(f)}{\partial f}\\
&= -D(f)\frac{\partial \phi_{d}(f)}{\partial f} \\
&= -D(f)w_{d}\label{eq:gradient_linear_discriminator}
\end{align}
where we obtain \eqref{eq:gradient_linear_discriminator} under the assumption that the discriminator is linear, i.e., $\phi_{d}(f)=w_{d}^{\top}f$, which is equivalent to the first one in \eqref{eq:gradient-dann-ss}.

\subsection{Gradient for {\dannssl}}
Let $\phi_{y}\,{:}\,\mathbb{R}^{\ndim}\,{\rightarrow}\,\mathbb{R}, y\,{=}\,1,...,\mathcal{Y}\,{+}\,1$ be a function that generates the exponent of classification distribution of {\dannssl}, i.e., $\overbar{C}(y) = \nicefrac{\exp(\phi_{y}(f))}{\sum_{y'=1}^{\mathcal{Y}+1}\exp(\phi_{y'}(f))}$. The gradient of adversarial loss in \eqref{eq:dann-semisup-adv} with respect to $f$ is written as follows:
\begin{equation}
\frac{\partial\log(1{-}\overbar{C}(\ncls{+}1))}{\partial f} = \frac{-1}{1{-}\overbar{C}(\ncls{+}1)}\frac{\partial\overbar{C}(\ncls{+}1)}{\partial f}\label{eq:grad-analysis-first}
\end{equation}
and the second term of RHS is written as
\begin{align}
\frac{\partial\overbar{C}(\ncls{+}1)}{\partial f} &=\frac{\phi'_{{\ncls}{+}1}\exp(\phi_{{\ncls}{+}1})}{\sum_{y'=1}^{\mathcal{Y}+1}\exp(\phi_{y'})} - \frac{\exp(\phi_{{\ncls}{+}1})\sum_{y'=1}^{\mathcal{Y}+1}\phi'_{y'}\exp(\phi_{y'})}{\{\sum_{y'=1}^{\mathcal{Y}+1}\exp(\phi_{y'})\}^2} \\
&= \phi'_{{\ncls}{+}1}\overbar{C}(\ncls{+}1) - \overbar{C}(\ncls{+}1)\sum_{y=1}^{\mathcal{Y}+1}\phi'_{y}\overbar{C}(y) \\
&= \phi'_{{\ncls}{+}1}\overbar{C}(\ncls{+}1)(1{-}\overbar{C}(\ncls{+}1)) - \overbar{C}(\ncls{+}1)\sum_{y=1}^{\mathcal{Y}}\phi'_{y}\overbar{C}(y)\label{eq:grad-analysis-second}
\end{align}
where $\phi'_{y} = \frac{\partial \phi_{y}(f)}{\partial f}$. Plugging \eqref{eq:grad-analysis-second} into \eqref{eq:grad-analysis-first} results in the following:
\begin{align}
\frac{\partial\log(1{-}\overbar{C}(\ncls{+}1))}{\partial f} &= -\phi'_{{\ncls}{+}1}\overbar{C}(\ncls{+}1) + \overbar{C}(\ncls{+}1)\sum_{y=1}^{\mathcal{Y}}\phi'_{y}\overbar{C}(y|\mathcal{Y}) \label{eq:grad-analysis-second-to-first} \\
&= -w_{\ncls{+}1}\overbar{C}(\ncls{+}1) + \overbar{C}(\ncls{+}1)\sum_{y=1}^{\mathcal{Y}}w_{y}\overbar{C}(y|\mathcal{Y}) \label{eq:grad-analysis-second-to-first-linear}
\end{align}
where we assume linear classifier and discriminator, $\phi_{y}(f) = w_{y}^{\top}f, y\,{=}\,1,...,N{+}1$ to derive \eqref{eq:grad-analysis-second-to-first-linear} from \eqref{eq:grad-analysis-second-to-first}.

% ------- MCD -------
\section{Relation to Maximum Classifier Discrepancy~\cite{Saito_2018_CVPR}}
\label{sec:supp_MCD}
Here we provide a detailed derivation of relation between our {\dannssl} and recently proposed Maximum Classifier Discrepancy (MCD) learning~\cite{Saito_2018_CVPR}, one of the consistency-based learning frameworks~\cite{laine2016temporal,tarvainen2017mean,french2017self}, presented in Section~\ref{sec:domadapt-semisup}.

\subsection{Maximum Classifier Discrepancy for Unsupervised Domain Adaptation}
We review the MCD learning framework for unsupervised domain adaptation. Similarly to the setting of the DANN, the MCD learning divides the classifier parameterized by deep neural networks into feature extractor ($f\,{:}\,\mathcal{X}{\to}\mathbb{R}^{\ndim}$) and classifiers built on top of feature extractor. Differently, it contains two (or more) classifiers $F_{i}\,{:}\,\mathbb{R}^{\ndim}{\rightarrow}(0,1)^{\mathcal{Y}}$ with no domain discriminator.

The learning proceeds as follows: First, two classifiers are trained (while fixing the feature extractor) to minimize the classification loss on the source domain while making maximally different prediction between classifiers on the target domain. Second, feature extractor is trained (while fixing classifiers) to minimize the classification loss on the source domain while making consistent prediction between classifiers on the target domain. The learning objective is written as follows:
\begin{gather}
\max_{F_{1},F_{2}} \mathbb{E}_{(x,y)\in\mathcal{X}_{S}\times\mathcal{Y}} \big[\log F_{1}(f,y) + \log F_{2}(f,y)\big] + \mathbb{E}_{x\in\mathcal{X}_{T}} d\left(F_{1}(f,\cdot), F_{2}(f,\cdot)\right)\\
\max_{f} \mathbb{E}_{(x,y)\in\mathcal{X}_{S}\times\mathcal{Y}} \big[\log F_{1}(f,y) + \log F_{2}(f,y)\big] - \mathbb{E}_{x\in\mathcal{X}_{T}} d\left(F_{1}(f,\cdot), F_{2}(f,\cdot)\right)
\end{gather}
The choice of discrepancy metric $d$ could be diverse and $L1$-distance $d(p_{1},p_{2})\,{=}\,\frac{1}{\ncls}\sum_{y=1}^{\ncls} |p_{1}(y)\,{-}\,p_{2}(y)|$ is used in~\cite{Saito_2018_CVPR}.

\subsection{Relation between {\dannssl} and Maximum Classifier Discrepancy~\cite{Saito_2018_CVPR}}
Now we derive the relation between {\dannssl} and MCD learning presented in Section~\ref{sec:domadapt-semisup} with more details. Following~\cite{Saito_2018_CVPR}, we define the two classification distributions:
\begin{equation}
p_{1}(y|x_{t}) \,{=}\, \overbar{C}(y|\mathcal{Y}),\, p_{2}(y|x_{t}) \,{=}\, \overbar{C}(y),\, y\,{\leq}\,\ncls{+}1
\end{equation}
Note that two classifiers $F_{1}$ and $F_{2}$ are both represented as $(\ncls{+}1)$-way classifier parameterization in {\dannssl}. Using KL divergence as discrepancy metric between two distributions, we obtain following discrepancy loss:
\begin{align}
-\text{KL}(p_{1}\Vert p_{2}) &= -\sum\nolimits_{y}^{\ncls+1}p_{1}(y)\log \frac{p_{1}(y)}{p_{2}(y)}\\
&= -\sum\nolimits_{y}^{\ncls}p_{1}(y)\log \frac{p_{1}(y)}{p_{2}(y)}\label{eq-supp:relation_to_MCD_step1}\\
&= -\sum\nolimits_{y}^{\ncls}p_{1}(y)\log \frac{1}{1{-}\overbar{C}(\ncls{+}1)}\label{eq-supp:relation_to_MCD_step2}\\
&=\log(1{-}\overbar{C}(\ncls{+}1))\label{eq-supp:relation_to_MCD}
\end{align}
where \eqref{eq-supp:relation_to_MCD_step1} is due to $\overbar{C}(\ncls{+}1|\mathcal{Y})\,{=}\,0$ and \eqref{eq-supp:relation_to_MCD_step2} is due to $\overbar{C}(y|\mathcal{Y})\,{=}\,\frac{\overbar{C}(y)}{1\,{-}\,\overbar{C}(\ncls{+}1)}$ for all $y\neq\ncls$. In other words, besides the specific choice of two classifiers and discrepancy kernels ($L1$-distance versus $\text{KL}$ divergence), two frameworks are indeed equivalent and thus are expected to have similar empirical performances as well. Empirical comparison of UDA methods including our proposed {\dannssl}, MCD~\cite{Saito_2018_CVPR}, as well as other consistency-based methods~\cite{laine2016temporal,tarvainen2017mean,french2017self} is left as a future work.

% ------- domadapt -------
\section{Unsupervised Model Selection}
\label{sec:supp-model-selection}
Model selection is an important component of unsupervised domain adaptation research since, we seldom have labeled examples from the target domain for validation due to its nature. Therefore, unsupervised model selection, i.e., model selection without using labeled examples from the target domain, is an essential component for any UDA method to be useful in practice. In this section, we introduce a variant of reverse validation~\cite{zhong2010cross,ganin2016domain}, the only unsupervised model selection method to our knowledge, and compare its effectiveness in comparison to our supervised model selection protocol using 5 images per output classes.

Reverse validation~\cite{zhong2010cross,ganin2016domain} is proposed to validate the performance of domain adaptation methods without using labeled examples from the target domain. The protocol is given as follows: 
\begin{tight_itemize}
\item{Train domain adaptation model (forward classifier) from source to target;}
\item{Train a ``reverse'' classifier from unlabeled target examples using pseudo labels predicted by the forward classifier;}
\item{Evaluate the performance of ``reverse'' classifier on labeled source examples.}
\end{tight_itemize}
The intuition is that if the forward classifier works well on the target examples, then the reverse classifier will also do well on the source domain, where one can have many labeled examples.

As the procedure introduces a new reverse classifier, the selection of classification method seems important. It is suggested from \cite{ganin2016domain} to use the same network architecture, possibly initialized from the same network parameters of forward classifier as reverse classifier. However, we find that this selection is not particularly attractive for the following reasons. Firstly, the reverse classifier, which is another deep neural network, is expensive and non-trivial to train, e.g., it may require additional hyperparameter tuning as two domains are not always symmetric. Secondly, deep networks are robust to noise and sometimes adding label noise improves the generalization performance of deep neural network~\cite{xie2016disturblabel}. These observations suggest limited correctness of the assumption of reverse validation whereby more accurate forward classifier leads to more accurate reverse classifier. For example, our experiment with office database shows that accuracies of reverse classifiers\footnote{For simplicity, we train a classifier of the same network configuration to forward classifier with self-labeled target domain examples, but without adaptation loss.} on labeled source examples with DANN and {\dannssl} as forward classifier on A$\rightarrow$W task are $66.21\%$ and $58.57\%$, respectively, while the performance of forward classifier on target examples are $72.33\%$ and $77.38\%$. Note that the performance of reverse classifier using the ground-truth labels as self-labeled target set is only $46.43\%$, which verifies the effectiveness of noisy labels in training deep neural network.

\begin{table*}[t]
\small
\begin{center}
\begin{tabular}{cccc ccccccccc}
\toprule
\multirow{2}{*}{ID} & \multirow{2}{*}{Persp.} & \multirow{2}{*}{Photo.} & \multirow{2}{*}{Feature} & \multicolumn{3}{c}{5/cls (sup.)} & \multicolumn{3}{c}{5-NN (unsup.)} & \multicolumn{3}{c}{mAP (unsup.)}\\ \cmidrule(l){5-7} \cmidrule(l){8-10} \cmidrule(l){11-13}
 & & & & Top-1 & Day & Night & Top-1 & Day & Night & Top-1 & Day & Night \\ \midrule
M1 & \multicolumn{3}{c}{Baseline (web only)} & 54.98 & 72.67 & 19.87 & \multicolumn{3}{c}{--} & \multicolumn{3}{c}{--} \\
M2 & \multicolumn{3}{c}{Supervised (web + SV)} & 98.63 & 98.92 & 98.05 & \multicolumn{3}{c}{--} & \multicolumn{3}{c}{--} \\ \midrule
M3 & AF & -- & -- & 59.73 & 75.78 & 27.87 & 58.88 & 75.27 & 26.35 & 58.89 & 75.64 & 25.64\\
M4 & KF & -- & -- & 61.55 & 77.98 & 28.92 & 60.87 & 76.70 & 29.45 & 60.47 & 76.56 & 28.52 \\
M5 & MKF & -- & -- & 64.30 & 78.62 & 35.87 & 61.63 & 75.53 & 34.04 & 64.37 & 78.67 & 35.99 \\
\midrule
M6 & -- & {CycleGAN} & -- & 64.32 & 77.01 & 39.12 & 60.92 & 73.55 & 35.87 & 61.25 & 73.95 & 36.02 \\
M7 & -- & {\accgan} & -- & 67.30 & 78.20 & 45.66 & 67.44 & 78.53 & 45.41 & 64.52 & 76.12 & 41.48 \\
M8 & MKF & CycleGAN & -- & 71.21 & 81.54 & 50.68 & 69.42 & 79.59 & 49.23 & 70.85 & 81.95 & 48.82 \\
M9 & MKF & {\accgan} & -- & 79.71 & 84.10 & 70.99 & 74.98 & 79.70 & 65.62 & 78.80 & 83.18 & 70.09 \\
\midrule
M10 & -- & -- & DANN & 60.40 & 75.56 & 30.31 & 58.15 & 73.97 & 26.74 & 60.05 & 75.52 & 29.32 \\
M11 & -- & -- & {\dannssl} & 75.83 & 76.73 & 74.05 & 75.01 & 76.53 & 71.99 & 75.40 & 76.51 & 73.19 \\
M12 & MKF & -- & {\dannssl} & 80.40 & 82.50 & 76.22 & 77.26 & 82.44 & 66.98 & 75.85 & 82.42 & 62.82 \\
M13 & -- & {\accgan} & {\dannssl} & 80.24 & 82.15 & 76.44 & 77.69 & 82.17 & 68.78 & 77.91 & 82.15 & 69.50 \\
M14 & MKF & {\accgan} & {\dannssl} & \textbf{84.20} & \textbf{85.77} & \textbf{81.10} & \textbf{83.78} & \textbf{85.54} & \textbf{80.27} & \textbf{83.82} & \textbf{85.56} & \textbf{80.37} \\
\bottomrule
\end{tabular}
\end{center}
\caption{Car recognition accuracy on surveillance images of CompCars dataset of our recognition system with different combinations of components evaluated by supervised and unsupervised model selection methods. We consider pixel-based (AF), keypoint-based (KF) and with mask (MKF) for perspective transformation, CycleGAN and attribute-conditioned CycleGAN, and DANN, {\dannssl} as variations.\label{tab:recog-full}}
\end{table*}

Instead, we propose few alternatives that are much simpler and more efficient to evaluate based on non-parameteric classifiers. We summarize our proposed unsupervised validation metrics below.
\begin{enumerate}
\item{k-nearest neighbor: we use k-nearest neighbor classifier using learned representation of forward model $f$ and predicted labels $C(f)$ (or $\overbar{C}(f|\mathcal{Y})$ for {\dannssl}) on target examples by forward model. The performance measure evaluated on labeled source data is given as follows:
\begin{equation}
\text{ACC}_{\text{kNN}} = \mathbb{E}_{(x,y)\in\mathcal{X}_{\text{S}}\times\mathcal{Y}_{\text{S}}}\textbf{1}{\{y=\arg\max_{\tilde{y}}\frac{1}{k}\sum_{\tilde{x}\in\text{kNN}(x)}C(f(\tilde{x}), \tilde{y})\}}
\end{equation}
We use $k=5$ for all our experiments.
}
\item{mAP: we use an average precision (AP) of labeled source examples with label-predicted target examples via forward classifier. The performance measure is given as follows:
\begin{equation}
{\text{mAP}} = \mathbb{E}_{(x,y)\in\mathcal{X}_{\text{S}}\times\mathcal{Y}_{\text{S}}} AP(x,y|\{x_{t},\arg\max_{y'} C(f(x_{t}), y')\}_{x_{t}\in\mathcal{X}_{\text{T}}})
\end{equation}
}
\end{enumerate}
The results with our proposed model selection methods are found in Table~\ref{tab:recog-full}. We observe that non-parameteric classifiers defined on learned representation can find models that are consistent with test set performance of the models chosen by supervised model selection method using 5 images per class. Although we find these unsupervised metrics effective, we also observe significant performance drop for some models selected by 5-NN or mAP (e.g., M6--M7, M12--M13). We believe that unsupervised model selection in deep domain adaptation is not yet solved and requires significant more investigation, both from empirical and theoretical perspectives, which is beyond the scope of our work and will leave them as a future work.

% ------- implementation -------
\section{Implementation Details}
We provide implementation details of individual components. All components are implemented in Torch~\cite{collobert:2011c}.

\subsection{Appearance Flow Estimation Networks}
\begin{figure*}[t]
\begin{center}
\includegraphics[width=0.7\linewidth]{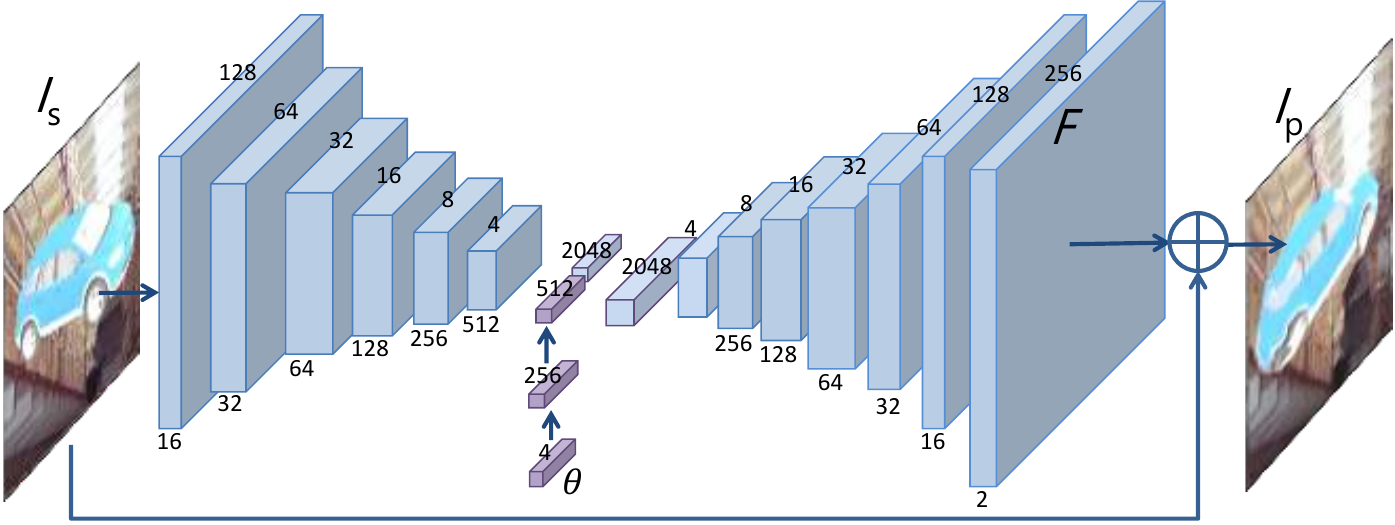}
\caption{AFNet architecture. AFNet receives source image $I_s$ and the target perspective $\theta$ (e.g., 4-dimensional one hot vector for elevation from $0^{\circ}{\sim}30^{\circ}$) as input and generates the flow field $F$ to synthesize image $I_p$ through bilinear sampling. \label{fig:AFNet}}
\end{center}
\end{figure*}

AFNet has an encoder-decoder structure, which is visualized in \Fig{fig:AFNet}. AFNet takes a source image and target viewpoint as input, where an image of size $256{\times}256$ is fed to a convolutional encoder to produce a $2048$-dimensional vector and it is concatenated with $512$-dimensional vector generated from the latent viewpoint code via viewpoint encoder. $2560$-dimensional concatenated vector is fed to decoder, which is constructed with fractionally-strided convolution layers, to generate flow representation of size $256{\times}256{\times}2$. Finally, a source image is warped via appearance flow based on bilinear sampling~\cite{jaderberg2015spatial,zhou2016view}\footnote{\url{https://github.com/qassemoquab/stnbhwd}} to predict a target image. All convolution layers use $3{\times}3$ filters, meanwhile filters of fractionally-strided convolution layers have size of $4{\times}4$. AFNet is trained using Adam optimizer~\cite{kingma2014adam} with the learning rate of $0.0003$ and batch size of $256$.

KFNet architecture is inherited from AFNet and shares the decoder architecture and viewpoint encoder. To accommodate sparse keypoints as the input, the entire image encoder is replaced by the keypoint encoder, consisting of two fully connected layers with $256$ and $2048$ output neurons, respectively. KFNet is trained to optimize \eqref{eqn:kf} with $\lambda\,{=}\,1$. Other hyperparameters such as the learning rate are the same as those used for AFNet training.

\subsection{Attribute-conditioned CycleGAN}
\begin{figure*}[htbp]
\begin{center}
\includegraphics[width=0.9\linewidth]{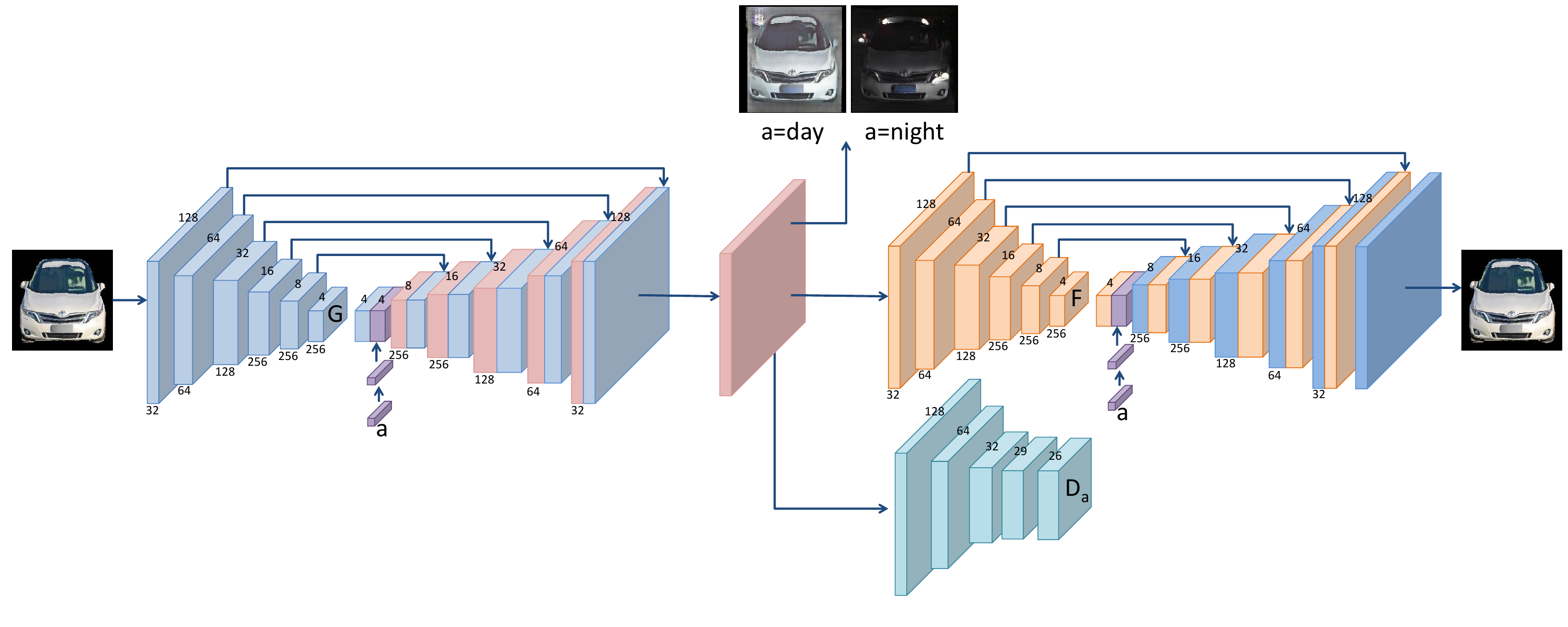}
\caption{Generator and discriminator network architectures of {\accgan}.\label{fig:cycleGAN_shared}}
\end{center}
\end{figure*}

The network architecture for generators and discriminators are illustrated in \Fig{fig:cycleGAN_shared}. The images of size $256{\times}256$ are used across input or output of generators and discriminators. UNet architecture~\cite{isola2017image} is used for both generators $G$ and $F$ while we feed the attribute code $a$ in the middle of the generator network. The $70{\times}70$ patchGAN discriminator~\cite{isola2017image} is used that generates $26{\times}26$-dimensional output for real/fake discrimination. The discriminator of conditional GAN~\cite{mirza2014conditional} is used where $D$ takes attribute code as an additional input to the real or generated images. One can consider a multi-way discriminator~\cite{taigman2016unsupervised,chen2016infogan} that discriminates not only between real or generated but also between different attribute configurations, but we didn't find it effective in our experiment.

We train using Adam optimizer with learning rate of $0.0002$ and the batch size of $32$ for all networks. In addition, we adopt two techniques from recent works to stabilize training procedure. For example, we replace the negative log likelihood objective of discriminator by a least square loss~\cite{mao2016multi,zhu2017unpaired}. Furthermore, we adopt historical buffer strategy~\cite{shrivastava2016learning} that updates the discriminator not only using generated images with the current generator but also with the generated images from the previous updates. We maintain an image buffer that stores the $1000$ previously generated images for each generator and randomly select $32$ images in the buffer to update discriminator.

\subsection{Domain Adversarial Neural Networks}
\label{sec:supp_implementation_details_dann}
The ImageNet pretrained ResNet-18~\cite{he2016deep}\footnote{\url{https://github.com/facebook/fb.resnet.torch/tree/master/pretrained}} fine-tuned on the CompCars web dataset is used as our baseline network. The dimension of the last fully-connected layer is $512$. The linear classifier ($512$ -- $431$) is used for all models. For discriminator, we try both linear ($512$ -- $1$) and MLP with different depth ($512$ -- $320{\times}d$ -- $1$, $d\,{=}\,1,...,4$) discriminators. The validation accuracy is given in Table~\ref{tab:discriminator_arch} and we decide to use 3-layer ($d\,{=}\,2$) MLP discriminator. Therefore, we employ linear discriminator ($512$ -- $432$) for our proposed {\dannssl}, while using MLP discriminator for standard DANN. We augment the classifier of the baseline network by adding one more column to construct the weight matrices for classifiers of {\dannssl}. The $432^{\text{nd}}$ column of the weight matrix is initialized by averaging the previous $431$ weight vectors, i.e., $w_{i,432}\,{=}\,\frac{1}{431}\sum_{k\,{=}\,1}^{431}w_{i,k}, i\,{=}\,1,...,512$.

For data preprocessing, we crop and scale web images into $256{\times}256$ using provided bounding boxes while maintaining the aspect ratio. Since they are already cropped, surveillance images are simply scaled into $256{\times}256$. We further crop an image of size $224{\times}224$ at random location of an image of size $256{\times}256$ with a random horizontal flip to feed to our feature extractor. All models are trained by updating the classifier/discriminator and CNN parameters in turn. Adam optimizer is used for training with the learning rate of $0.00001$, which is equivalent to the final learning rate of the fine-tuned model on CompCars web dataset. In addition to $\lambda$ in \eqref{eq:dann} and \eqref{eq:dann-semisup-adv} that balances classification loss and domain adversarial loss for updating parameters of feature extractor, we also tune learning rates of classifier and discriminator separately. Specifically, we augment \eqref{eq:dann-disc} and \eqref{eq:dann-semisup} as follows:
\begin{align}
\max_{\theta_{d}}&\{\mathcal{L}_{\text{D}} \,{=}\,\mathbb{E}_{\mathcal{X}_{\text{S}}} \log (1{-}D(f)) + \beta\mathbb{E}_{\mathcal{X}_{\text{T}}} \log D(f)\} \label{eq:dann-disc-param}\\
\max_{\theta_{c}}& \{\overbar{\mathcal{L}}_{\text{C}}\,{=}\,\mathbb{E}_{\mathcal{X}_{\text{S}}}\log \overbar{C}(y) + \beta\mathbb{E}_{\mathcal{X}_{\text{T}}}\log \overbar{C}(\ncls{+}1)\}\label{eq:dann-semisup-param}
\end{align}
We apply regularization coefficient $\beta$ to loss induced by the target examples. When $\beta\,{=}\,1$, it becomes equivalent to that of \cite{ganin2016domain}.\footnote{We set $\beta\,{=}\,1$ for experiments on office database in Section~\ref{supp:benchmark-office} following the implementation by~\cite{ganin2016domain} to inherit most of the training protocol such as hyperparameter setting.} In these experiments, we find that $\beta\,{=}\,\frac{1}{\ncls}$ is a good starting point for hyperparameter search of {\dannssl}, where $\ncls\,{=}\,431$ is the number of classes and we finally fix $\beta \,{=}\, 0.001$ for models used in experiments on CompCars dataset. Due to small $\beta$, we increase $\lambda$ for {\dannssl} to backpropagate sufficient amount of gradient from adversarial loss. We also tune $\beta$ for DANN from $\{100,10,1,0.1,0.01,0.001\}$, but we don't observe significant performance difference. As a result we fix $\beta\,{=}\,1$ for DANN. The optimal setting of other hyperparameters are reported in Table~\ref{tab:hyperparams_compcars}.

\begin{table*}[t]
\small
\begin{center}
\begin{tabular}{ lccccc}
\toprule 
& linear & MLP with $d\,{=}\,1$ & MLP with $d\,{=}\,2$ & MLP with $d\,{=}\,3$ & MLP with $d\,{=}\,4$ \\
\midrule
Accuracy & 58.40{$\pm$0.59} & 59.11{$\pm$0.79} & 60.01{$\pm$0.74} & 59.23{$\pm$0.66} & 59.45{$\pm$0.68} \\
\bottomrule
\end{tabular}
\end{center}
\caption{Car recognition accuracy on SV validation set of CompCars of DANNs with different discriminator architectures.\label{tab:discriminator_arch}}
\end{table*}

\begin{table}[t]
\small
\begin{center}
\begin{tabular}{cc}
\toprule 
 {DANN} & {\dannssl} \\
\midrule
 $\beta\,{=}\,1,\lambda\,{=}\,0.01$ & {$\beta\,{=}\,0.001,\lambda\,{=}\,100$} \\
\bottomrule
\end{tabular}
\end{center}
\caption{Optimal hyperparameters on CompCars dataset.\label{tab:hyperparams_compcars}} 
\end{table}

% ------- modsel -------
\section{Details on Section~\ref{sec:exp-feat}: ``Evaluation on UDA Benchmark''}
\label{supp:benchmark_details}
We provide details for the evaluation of DANNs on standard UDA benchmarks. As presented in Section~\ref{sec:exp-feat}, we evaluate on four tasks of digits and traffic sign recognition problems~\cite{ganin2016domain} and six tasks of office object recognition problems~\cite{saenko2010adapting}. The details, such as task description or experimental results, of individual experiments are discussed below.

\subsection{Digits and Traffic Signs}
\label{supp:benchmark-digits}
\subsubsection{Task Description}
We start the section by task description and model selection. We note that supervised model selection using a subset of labeled target examples is used for this experiment inspired by~\cite{bousmalis2016domain}.
\begin{enumerate}
\item{MNIST$\rightarrow$MNIST-M: MNIST-M is a variation of MNIST with color-transformed foreground digits over natural images in the background. Following~\cite{haeusser2017associative}, we augment source data by inverting pixel-values from 0 to 255 and vice versa, thus doubling the volume of source data. Overall, $120K (=60K\,{\times}\,2)$ labeled source images, $50K$ unlabeled target images for training, $1,000$ labeled target images for validation, $9,001$ labeled target images for testing are used.}
\item{Synthetic Digits$\rightarrow$SVHN: Synthesized digits~\cite{ganin2016domain} are used as labeled training examples to recognize digits in street view house number dataset (SVHN)~\cite{netzer2011reading}. Unlike other works, we use extra unlabeled images of SVHN dataset to train adaptation models. Overall, $479,400$ labeled source images, $581,131$ unlabeled target images for training, $1,000$ labeled target images for validation, $26,032$ labeled target images for testing are used.}
\item{SVHN$\rightarrow$MNIST: SVHN is used as a source and MNIST is used as a target. Overall, $73,257$ labeled source images, $50K$ unlabeled target images for training, $1,000$ labeled target images for validation, $10K$ labeled target images for testing are used.}
\item{MNIST$\rightarrow$SVHN: MNIST is used as a source and SVHN is used as a target. Overall, $50K$ labeled source images, $73,257$ unlabeled target images for training, $1,000$ labeled target images for validation, $26,032$ labeled target images for testing are used.}
\item{Synthetic Signs$\rightarrow$GTSRB: In this task we recognize traffic signs from german traffic sign recognition benchmark (GTSRB)~\cite{stallkamp2011german} by adapting from labeled synthesized images~\cite{moiseev2013evaluation}. In total, $90K$ labeled source images, $35K$ unlabeled target images for training, $430$ labeled target images for validation, $12,569$ labeled target images for testing are used. Unlike other tasks with 10-way classification using $32\,{\times}\,32$ images as input, this task is 43-way classification and input images are of size $40\,{\times}\,40$.}
\end{enumerate}
For all tasks, we apply the same data preprocessing of channel-wise mean and standard deviation normalization per example~\cite{haeusser2017associative}, i.e., 
\begin{equation}
\tilde{x}_{i,j,c} = \frac{(x_{i,j,c} -  \bar{x}_{c})}{\hat{x}_{c}}
\end{equation}
where 
\begin{gather}
\bar{x}_{c} = \frac{1}{w\times h}\sum_{i=1}^{w}\sum_{j=1}^{h} x_{i,j,c},\\
\hat{x}_{c} = \sqrt{\frac{1}{(w\times h)-1}\sum_{i=1}^{w}\sum_{j=1}^{h} (x_{i,j,c}-\bar{x})^2}.
\end{gather}

\begin{figure*}[t]
\small
\begin{center}
\subfigure[M$\rightarrow$MM]{\includegraphics[width=0.23\textwidth]{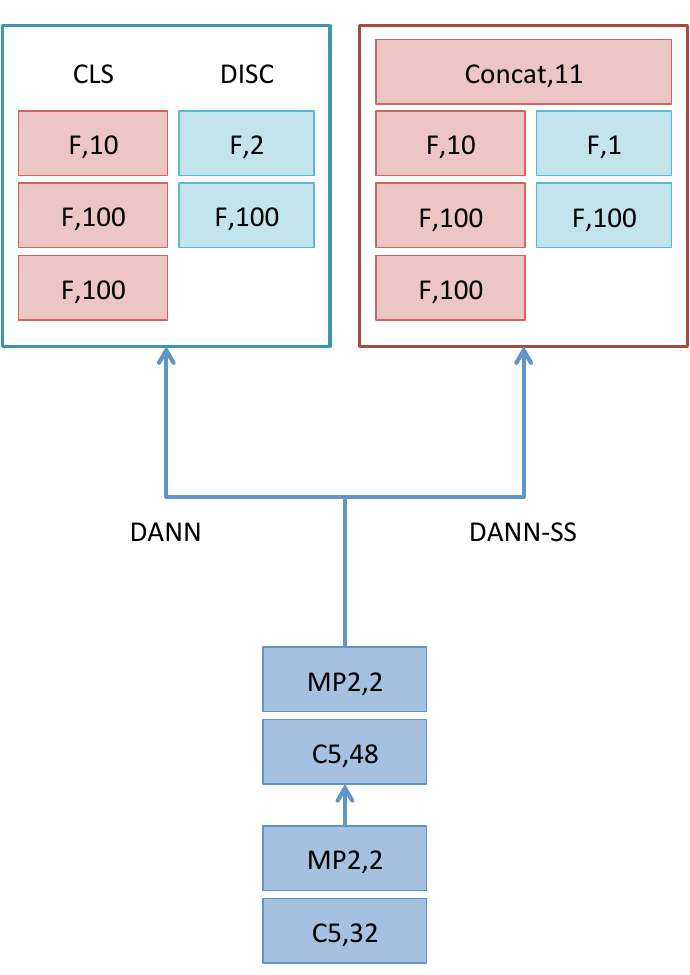}\hspace{0.02in}}
\subfigure[S$\rightarrow$S, S$\rightarrow$M, M$\rightarrow$S]{\includegraphics[width=0.23\textwidth]{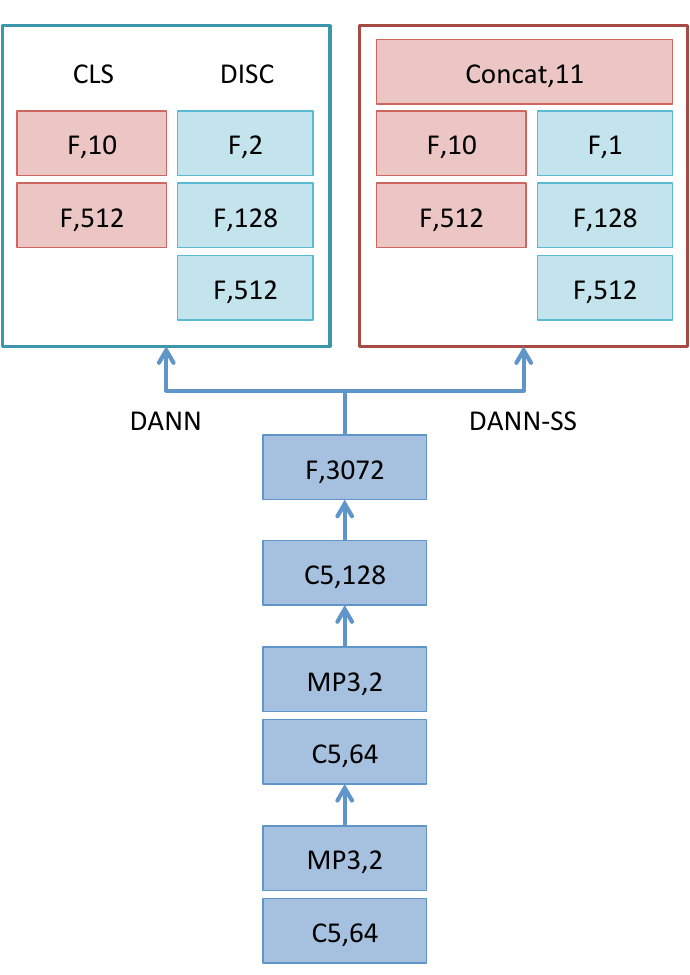}\hspace{0.02in}}
\subfigure[S$\rightarrow$G]{\includegraphics[width=0.23\textwidth]{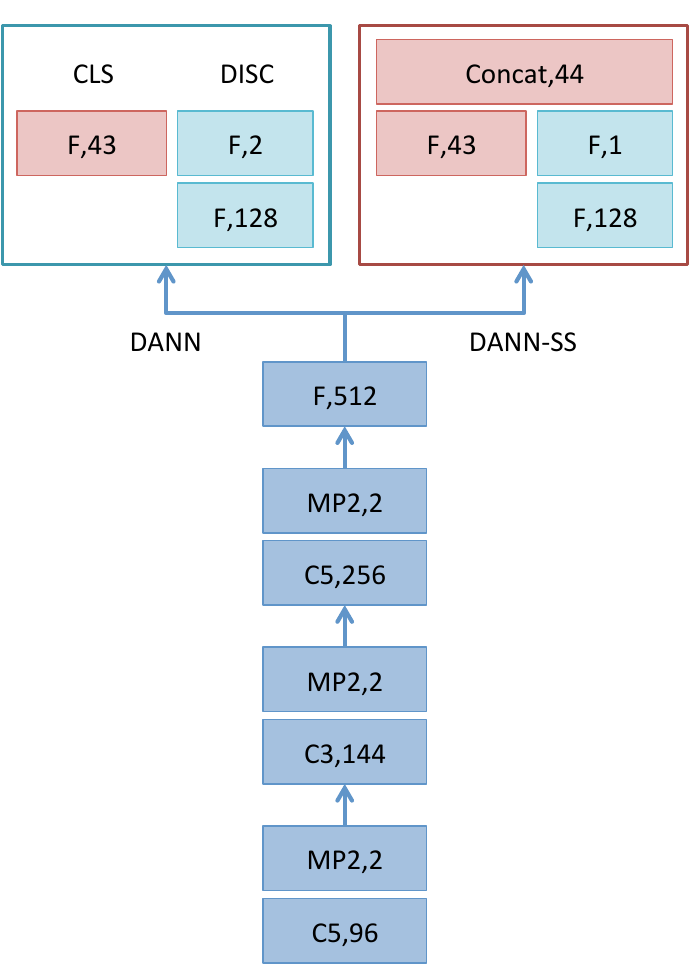}\hspace{0.02in}}
\subfigure[deep (all)]{\includegraphics[width=0.23\textwidth]{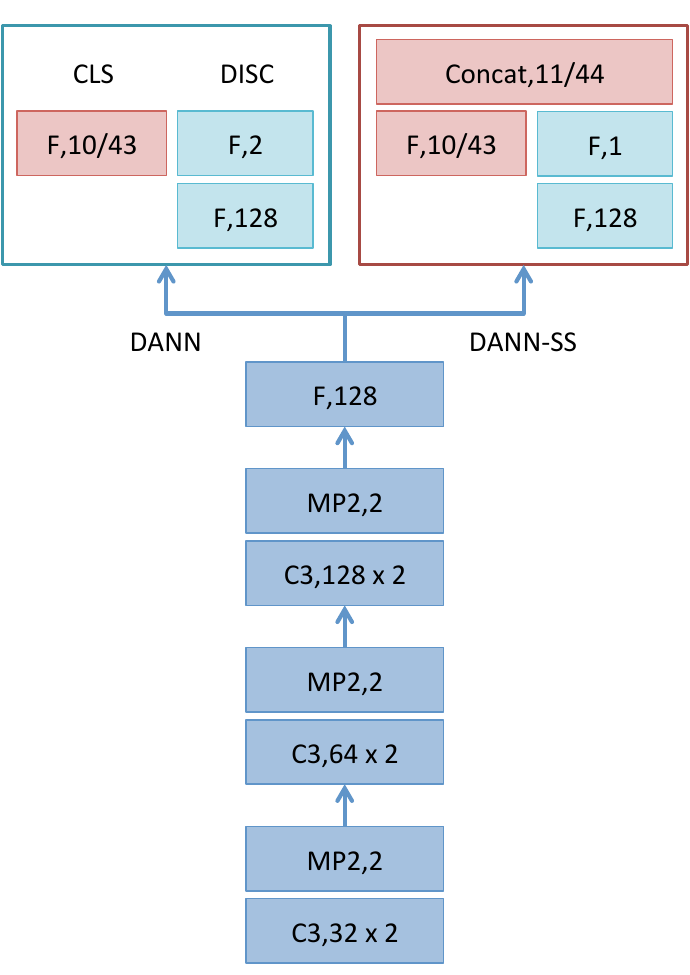}}
\end{center}
\caption{(a-c) Shallow~\cite{ganin2016domain} and (d) deep~\cite{haeusser2017associative} network architectures for digit and traffic sign adaptation tasks. Three different shallow architectures are used for different tasks following~\cite{ganin2016domain}. ReLU activation is applied followed by convolutional and fully-connected layers except for the last fully-connected layer connected to classifier or discriminator.\label{fig:netarch-benchmark}}
\end{figure*}

We experiment with shallow (2$\sim$3 convolution layers)~\cite{ganin2016domain} and deep (6 convolution layers)~\cite{haeusser2017associative} network architectures as described in \Fig{fig:netarch-benchmark}. The shallow network architectures are inspired by~\cite{ganin2016domain} and share the same convolution and pooling architecture, but the classifier and discriminator architectures are slightly different. Similarly, convolution and pooling architecture of deep network is the same as that of~\cite{haeusser2017associative} but classifier and discriminator are of our own design.

\subsubsection{Results}
The summary results are provided in Table~\ref{tab:dann-comparison-soa}. We train 10 models with different random seeds per method and task and report the mean test set error and standard error. When there is a tie in validation performance between models with different sets of hyperparameter or at different training epochs, which happens quite frequently since we are using small number of validation examples, we report the average test set performance of the models. The proposed {\dannssl} significantly improves the performance upon standard DANN on most tasks with both shallow and deep network architectures, achieving state-of-the-art results on 4 out of 5 tasks.

\begin{table*}[t]
\begin{center}
\small
\begin{tabular}{lccccccc}
\toprule
Method & $\#$ val. set & network & M$\rightarrow$MM & S$\rightarrow$S & S$\rightarrow$M & M$\rightarrow$S & S$\rightarrow$G \\ \midrule
RevGrad~\cite{ganin2016domain} & $0$ & shallow & 76.67 & 91.09 & 73.85 & -- & 88.65 \\ 
DSN~\cite{bousmalis2016domain} & $1000/430$ & shallow & 83.2 & 91.2 & 82.7 & -- & 93.1 \\ 
ADA~\cite{haeusser2017associative} & -- & deep & 89.53 & 91.86 & {\color{red}\textbf{97.6}} & -- & 97.66 \\
\midrule
source only & \multirow{3}{*}{$1000/430$} & \multirow{3}{*}{shallow} & 68.28{\tiny${\pm0.29}$} & 87.22{\tiny${\pm0.18}$} & 68.39{\tiny${\pm0.79}$} & 59.80{\tiny${\pm0.57}$} & 95.63{\tiny${\pm0.13}$} \\
DANN & & & 88.62{\tiny${\pm0.29}$} & 88.07{\tiny${\pm0.16}$} & 92.34{\tiny${\pm0.88}$} & 75.48{\tiny${\pm2.10}$} & 97.33{\tiny${\pm0.10}$} \\
{\dannssl} & & & \textbf{90.41}{\tiny${\pm0.20}$} & \textbf{93.32}{\tiny${\pm0.12}$} & \textbf{94.15}{\tiny${\pm1.42}$} & \textbf{82.96}{\tiny${\pm0.90}$} & \textbf{98.47}{\tiny${\pm0.09}$} \\
\midrule
source only & \multirow{3}{*}{$1000/430$} & \multirow{3}{*}{deep} & 67.90{\tiny${\pm0.95}$} & 87.05{\tiny${\pm0.22}$} & 63.74{\tiny${\pm0.68}$} & 62.44{\tiny${\pm0.52}$} & 94.53{\tiny${\pm0.14}$} \\
DANN & & & \textbf{98.00}{\tiny${\pm0.07}$} & 92.24{\tiny${\pm0.13}$} & 88.70{\tiny${\pm0.33}$} & 82.30{\tiny${\pm1.15}$} & 97.38{\tiny${\pm0.13}$} \\
{\dannssl} & & & {\color{red}\textbf{98.03}}{\tiny${\pm0.06}$} & {\color{red}\textbf{94.47}}{\tiny${\pm0.06}$} & {\color{red}\textbf{96.23}}{\tiny${\pm0.14}$} & {\color{red}\textbf{87.48}}{\tiny${\pm1.31}$} & {\color{red}\textbf{98.70}}{\tiny${\pm0.06}$} \\
\bottomrule
\end{tabular}
\end{center}
\caption{Evaluation on digit and traffic sign adaptation tasks, such as MNIST~\cite{lecun1998gradient} to MNIST-M~\cite{ganin2016domain} (M$\rightarrow$MM), Synthetic Digits~\cite{ganin2016domain} to SVHN (S$\rightarrow$S), SVHN to MNIST (S$\rightarrow$M), MNIST to SVHN (M$\rightarrow$S), or Synthetic Signs~\cite{moiseev2013evaluation} to GTSRB~\cite{stallkamp2011german} (S$\rightarrow$G). Experiments are executed for 10 times with different random seeds and mean test set accuracy and standard error are reported. For each network architecture, the best performers and the ones within standard error are bold-faced. Finally, the best performers across different architectures are colored in red.\label{tab:dann-comparison-soa}} 
\end{table*}

\subsection{Office Database}
\label{supp:benchmark-office}
\subsubsection{Task Description}
The office database~\cite{saenko2010adapting} is composed of three datasets, such as Amazon, Webcam, or DSLR, where each dataset contains images of 31 object categories from different sources. The number of images for each dataset is $2817$, $795$ and $498$, respectively. Individual dataset is considered as one domain and six adaptation tasks are experimented in total. We note that the office database is not particularly suitable to demonstrate the effectiveness of our proposed joint pixel and feature-level adaptation framework since there is no obvious way to inject pixel-level insights, such as 3D shape or lighting variations. In addition, as discussed in~\cite{bousmalis2016domain}, the dataset might be limited as there exists considerable amount of high-level variations such as label noise and the number of examples for training deep adaptation networks is not sufficient.

Nevertheless, the dataset is still useful to demonstrate the effectiveness of our proposed feature-level DA methods, such as {\dannssl}. We follow the training protocol of \cite{ganin2016domain}, where ImageNet-pretrained AlexNet~\cite{krizhevsky2012imagenet} is used to initialize the network parameters while the last fully-connected layer (4096 -- 1000) is replaced into shared bottleneck layer (4096 -- 256) followed by classifier (256 -- 31) and discriminator (256 -- 1024 -- 1024 -- 1). We also performed the same experiments with ImageNet-pretrained ResNet-50~\cite{he2016deep} following the protocol of~\cite{DBLP:journals/corr/LongCWJ17}. We use relatively shallower network architecture for classifier and discriminator, where we first replace the last fully-connected layer (2048 -- 1000) into shared bottleneck layer (2048 -- 256) followed by classifier (256 -- 31) and discriminator (256 -- 256 -- 1). For {\dannssl}, the output of classifier and discriminator are concatenated to form a unified classifier.

We optimize networks using momentum SGD with ``inv'' learning rate decay policy of Caffe~\cite{jia2014caffe}. We evaluate on the fully transductive setting~\cite{ganin2016domain,long2016unsupervised}, where all source and target examples are used for the training of deep networks. 

\begin{table*}[t]
\begin{center}
\small
\begin{tabular}{lcccccccc}
\toprule
\multicolumn{9}{c}{AlexNet} \\ \midrule
Method & Val. & A$\rightarrow$W & D$\rightarrow$W & W$\rightarrow$D & A$\rightarrow$D & D$\rightarrow$A & W$\rightarrow$A & Avg \\ \midrule
RevGrad~\cite{ganin2016domain} & RV & 73.0 & 96.4 & 99.2 & -- & -- & -- \\ 
RTN~\cite{long2016unsupervised} & sup-1 & 73.3 & 96.8 & 99.6 & 71.0 & 50.5 & 51.0 & 73.7 \\ 
CDAN-RM~\cite{DBLP:journals/corr/LongCWJ17} & IWCV & 77.9{\tiny${\pm0.3}$} & 96.9{\tiny${\pm0.2}$} & 100{\tiny${\pm0.0}$} & 74.6{\tiny${\pm0.2}$} & 55.1{\tiny${\pm0.3}$} & 57.5{\tiny${\pm0.4}$} & 77.0 \\
CDAN-M~\cite{DBLP:journals/corr/LongCWJ17} & IWCV & 77.6{\tiny${\pm0.2}$} & 97.2{\tiny${\pm0.1}$} & 100{\tiny${\pm0.0}$} & 73.0{\tiny${\pm0.1}$} & 57.3{\tiny${\pm0.2}$} & 56.1{\tiny${\pm0.3}$} & 76.9 \\
\midrule
\multirow{4}{*}{DANN} & 5-NN & 72.10{\tiny${\pm0.70}$} & 96.29{\tiny${\pm0.06}$} & 99.45{\tiny${\pm0.05}$} & 70.97{\tiny${\pm0.49}$} & 51.06{\tiny${\pm0.41}$} & 50.83{\tiny${\pm0.37}$} & 73.45 \\
& mAP & 72.33{\tiny${\pm0.61}$} & 96.43{\tiny${\pm0.11}$} & 99.76{\tiny${\pm0.04}$} & 70.96{\tiny${\pm0.42}$} & 51.33{\tiny${\pm0.34}$} & 51.23{\tiny${\pm0.49}$} & 73.67 \\
& sup-1 & 72.41{\tiny${\pm0.70}$} & 96.42{\tiny${\pm0.12}$} & 99.54{\tiny${\pm0.09}$} & 70.66{\tiny${\pm0.74}$} & 50.95{\tiny${\pm0.33}$} & 50.74{\tiny${\pm0.39}$} & 73.45 \\
& oracle & 73.64{\tiny${\pm0.51}$} & 96.86{\tiny${\pm0.10}$} & 99.92{\tiny${\pm0.04}$} & 72.09{\tiny${\pm0.55}$} & 51.98{\tiny${\pm0.17}$} & 51.91{\tiny${\pm0.32}$} & 74.40 \\
\cmidrule{2-9}
\multirow{4}{*}{\dannssl} & 5-NN & 77.23{\tiny${\pm1.37}$} & 96.87{\tiny${\pm0.03}$} & 99.56{\tiny${\pm0.12}$} & 74.10{\tiny${\pm0.93}$} & 59.23{\tiny${\pm0.62}$} & 57.89{\tiny${\pm0.81}$} & 77.48 \\
& mAP & 77.38{\tiny${\pm1.32}$} & 97.11{\tiny${\pm0.05}$} & 99.60{\tiny${\pm0.06}$} & 74.10{\tiny${\pm0.94}$} & 59.53{\tiny${\pm0.68}$} & 57.83{\tiny${\pm0.85}$} & 77.59 \\
& sup-1 & 77.31{\tiny${\pm1.52}$} & 97.00{\tiny${\pm0.07}$} & 99.68{\tiny${\pm0.13}$} & 73.69{\tiny${\pm1.00}$} & 58.79{\tiny${\pm0.80}$} & 57.31{\tiny${\pm0.89}$} & 77.30 \\
& oracle & 78.09{\tiny${\pm1.46}$} & 97.28{\tiny${\pm0.03}$} & 99.88{\tiny${\pm0.11}$} & 74.58{\tiny${\pm0.88}$} & 59.70{\tiny${\pm0.68}$} & 58.20{\tiny${\pm0.80}$} & 77.95 \\
\bottomrule
\end{tabular}
\end{center}
\caption{Evaluation on six adaptation tasks of Office benchmark using AlexNet. For each model and task, we report four numbers using different model selection mechanisms such as (first row) 5-NN classifier or (second row) mAP for reverse validation (RV) on source data, (third row) one labeled target example per class, or (fourth row) oracle selection via test set accuracy, which serves as an upper bound to aforementioned validation methods. All experiments are conducted 5 times with different random seeds and the mean accuracy and standard error are reported. \label{tab:office_benchmark_alexnet}}
\end{table*}

\begin{table*}[htbp]
\begin{center}
\small
\begin{tabular}{lcccccccc}
\toprule
\multicolumn{9}{c}{ResNet-50} \\ \midrule
Method & Val. & A$\rightarrow$W & D$\rightarrow$W & W$\rightarrow$D & A$\rightarrow$D & D$\rightarrow$A & W$\rightarrow$A & Avg \\ \midrule
RevGrad~\cite{DBLP:journals/corr/LongCWJ17} & IWCV & 82.0{\tiny${\pm0.4}$} & 96.9{\tiny${\pm0.2}$} & 99.1{\tiny${\pm0.1}$} & 79.7{\tiny${\pm0.4}$} & 68.2{\tiny${\pm0.4}$} & 67.4{\tiny${\pm0.5}$} & 82.2 \\
CDAN-RM~\cite{DBLP:journals/corr/LongCWJ17} & IWCV & 93.0{\tiny${\pm0.2}$} & 98.4{\tiny${\pm0.2}$} & 100{\tiny${\pm0.0}$} & 89.2{\tiny${\pm0.3}$} & 70.2{\tiny${\pm0.4}$} & 69.4{\tiny${\pm0.4}$} & 86.7 \\
CDAN-M~\cite{DBLP:journals/corr/LongCWJ17} & IWCV & 93.1{\tiny${\pm0.1}$} & 98.6{\tiny${\pm0.1}$} & 100{\tiny${\pm0.0}$} & 93.4{\tiny${\pm0.2}$} & 71.0{\tiny${\pm0.3}$} & 70.3{\tiny${\pm0.3}$} & 87.7 \\
\midrule
\multirow{4}{*}{DANN} & 5-NN & 86.29{\tiny${\pm0.28}$} & 96.95{\tiny${\pm0.10}$} & 98.01{\tiny${\pm0.12}$} & 83.99{\tiny${\pm0.45}$} & 66.58{\tiny${\pm0.40}$} & 67.08{\tiny${\pm0.12}$} & 83.15 \\
& mAP & 86.42{\tiny${\pm0.34}$} & 96.81{\tiny${\pm0.28}$} & 97.91{\tiny${\pm0.20}$} & 84.10{\tiny${\pm0.51}$} & 67.73{\tiny${\pm0.61}$} & 67.10{\tiny${\pm0.25}$} & 83.35 \\
& sup-1 & 85.97{\tiny${\pm0.51}$} & 96.87{\tiny${\pm0.19}$} & 97.94{\tiny${\pm0.14}$} & 84.12{\tiny${\pm0.50}$} & 67.63{\tiny${\pm0.73}$} & 66.78{\tiny${\pm0.33}$} & 83.22 \\
& oracle & 86.97{\tiny${\pm0.24}$} & 97.84{\tiny${\pm0.16}$} & 99.00{\tiny${\pm0.06}$} & 85.50{\tiny${\pm0.38}$} & 68.65{\tiny${\pm0.58}$} & 67.67{\tiny${\pm0.09}$} & 84.27 \\
\cmidrule{2-9}
\multirow{4}{*}{\dannssl} & 5-NN & 91.47{\tiny${\pm0.32}$} & 98.19{\tiny${\pm0.05}$} & 99.43{\tiny${\pm0.02}$} & 89.32{\tiny${\pm0.65}$} & 69.59{\tiny${\pm0.21}$} & 69.09{\tiny${\pm0.16}$} & 86.18 \\
& mAP & 91.47{\tiny${\pm0.32}$} & 98.26{\tiny${\pm0.11}$} & 99.52{\tiny${\pm0.04}$} & 89.28{\tiny${\pm0.61}$} & 70.11{\tiny${\pm0.17}$} & 69.34{\tiny${\pm0.21}$} & 86.33 \\
& sup-1 & 91.35{\tiny${\pm0.36}$} & 98.24{\tiny${\pm0.07}$} & 99.48{\tiny${\pm0.10}$} & 89.94{\tiny${\pm0.41}$} & 69.63{\tiny${\pm0.40}$} & 68.76{\tiny${\pm0.40}$} & 86.23 \\
& oracle & 92.20{\tiny${\pm0.26}$} & 98.47{\tiny${\pm0.04}$} & 99.60{\tiny${\pm0.00}$} & 90.64{\tiny${\pm0.20}$} & 70.64{\tiny${\pm0.19}$} & 69.70{\tiny${\pm0.22}$} & 86.88 \\
\bottomrule
\end{tabular}
\end{center}
\caption{Evaluation on six adaptation tasks of Office benchmark using ResNet-50. The same experimental protocol is employed to that using AlexNet. We also transfer the hyperparameters for each task from experiments usign AlexNet except that early stopping is done with respective model selection metrics.\label{tab:office_benchmark_resnet}}
\end{table*}

\subsubsection{{\dannssl} with Reverse Gradient}
To reduce an effort of additional hyperparameter search, we extend our proposed joint parameterization of classifier and discriminator for unsupervised domain adaptation to reverse gradient~\cite{ganin2016domain}, a pioneering method of domain adversarial neural network. The loss formulation is similar to that of standard DANN in \eqref{eq:dann} with a slight modification as follows:
\begin{align}
&\max_{\theta_{f}}\{\mathcal{L}_{\text{F}} \,{=}\,\mathcal{L}_{\text{C}} \,{-}\, \lambda\mathcal{L}_{\text{D}} =\mathcal{L}_{\text{C}} \,{-}\, \lambda\{\mathbb{E}_{\mathcal{X}_{\text{S}}} \log (1{-}D(f(x))) + \mathbb{E}_{\mathcal{X}_{\text{T}}} \log D(f(x))\}\} \label{eq:revgrad}
\end{align}
The losses for classifier and discriminator remain the same as in \eqref{eq:dann-cls} and \eqref{eq:dann-disc}. The negative sign on the adversarial loss in \eqref{eq:revgrad} amounts to reversing (and scaling with $\lambda$) the gradient before further backpropagating through $f$. This allows the entire network including classifier and discriminator as well as feature extractor to be trained end-to-end without alternating update. Besides the negative sign, we also notice that there is an additional source-to-target confusion term, $-\mathbb{E}_{\mathcal{X}_{\text{S}}}\log (1-D(f(x)))$, which we find playing an important role in this experiment. Inspired by our analysis, we use the following formulations of DANN and {\dannssl} in this experiment:
\begin{align}
&\max_{\theta_{f}} \{\mathcal{L}_{\text{F}} =\mathcal{L}_{\text{C}} + \lambda\{\mathbb{E}_{\mathcal{X}_{\text{S}}} \log D(f(x)) + \mathbb{E}_{\mathcal{X}_{\text{T}}} \log (1\,{-}\,D(f(x)))\}\} \label{eq:dann-revgrad}\\
&\max_{\theta_{f}} \{\widetilde{\mathcal{L}}_{\text{F}}=\mathbb{E}_{\mathcal{X}_{\text{S}}}\log \widetilde{C}(y|\mathcal{Y})+ \lambda\{\mathbb{E}_{\mathcal{X}_{\text{S}}}\log \widetilde{C}(\ncls{+}1) + \mathbb{E}_{\mathcal{X}_{\text{T}}}\log (1{-}\widetilde{C}(\ncls{+}1))\}\} \label{eq:dann-semisup-revgrad}
\end{align}
Note that instead of having $-\mathcal{L}_{\text{D}}$ we define adversarial losses by flipping source and target labels. As a result, model components are still trained alternatively between classifiers and feature extractor. Nonetheless, this allows us to transfer most of the hyperparameters from RevGrad implementation~\cite{ganin2016domain}\footnote{\url{https://github.com/ddtm/caffe/tree/grl}} including those related to SGD such as learning rate or its scheduler. We perform few hyperparameter searches for $\lambda$ starting from $0.1$ as suggested by \cite{ganin2016domain}.

\subsubsection{Results}
\label{supp:benchmark-office-discussion}
The mean accuracy and standard error of the standard DANN and {\dannssl} models trained with 5 different random seeds are reported in Table~\ref{tab:office_benchmark_alexnet} and \ref{tab:office_benchmark_resnet} using AlexNet and ResNet-50 as base networks, respectively. The proposed {\dannssl} improves upon the standard DANN by a significant margin. Moreover, the model demonstrates comparable performance to the state-of-the-art method~\cite{DBLP:journals/corr/LongCWJ17} on both experiments using AlexNet and ResNet-50 as backbone CNNs.

% ------- cyclegan -------
\begin{table}[t]
\small
\begin{center}
\begin{tabular}{cccc ccccccccc}
\toprule
{ID} & {Persp.} & {Photo.} & {Feature} & Top-1 & Day & Night \\
\midrule
\multirow{2}{*}{M7} & \multirow{2}{*}{--} & {\accgan} (shared) & \multirow{2}{*}{--} & 67.30 & 78.20 & 45.66 \\
 &  & {\accgan} (unshared) &  & 70.03 & 78.81 & 52.60 \\
 \midrule
\multirow{2}{*}{M9} & \multirow{2}{*}{MKF} & {\accgan} (shared) & \multirow{2}{*}{--} & 79.71 & 84.10 & 70.99 \\
 &  & {\accgan} (unshared) &  & 77.75 & 82.76 & 67.79 \\
 \midrule
\multirow{2}{*}{M14} & \multirow{2}{*}{MKF} & {\accgan} (shared) & \multirow{2}{*}{\dannssl} & 84.20 & 85.77 & 81.10 \\
 &  & {\accgan} (unshared) &  & \textbf{84.38} & \textbf{85.81} & \textbf{81.56} \\
\bottomrule
\end{tabular}
\end{center}
\caption{Comparison between {\accgan}s with shared and unshared parameters across generators and discriminators for car model recognition accuracy on CompCars Surveillance dataset. \label{tab:recog_supp}}
\end{table}

\section{{\accgan} with Unshared Parameters}
\label{supp:cyclegan}
\begin{figure}[htbp]
\begin{center}
\subfigure[Attribute-conditioned CycleGAN]{\includegraphics[width=0.9\linewidth]{attrCycleGAN.pdf}\label{fig:cycleGAN_shared2}}\hspace{0.08in}
\subfigure[Attribute-conditioned CycleGAN with unshared parameters]{\includegraphics[width=0.9\linewidth]{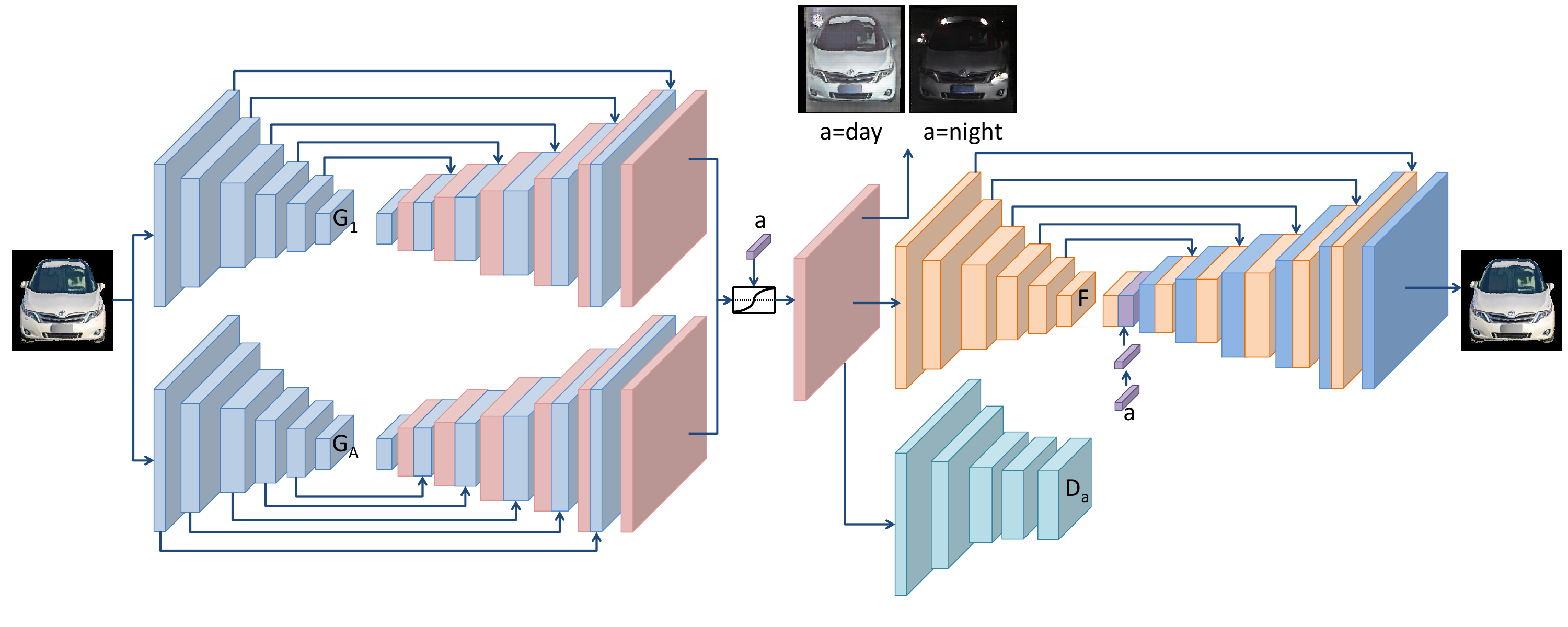}\label{fig:cycleGAN_unshared}}
\caption{Networks architecture comparisons between {\accgan}s with shared and unshared parameters across generators and discriminators.\label{fig:cycleGAN}}
\end{center}
\end{figure}

As we have small number of attribute configurations for lighting (e.g., day and night), it is affordable to use generator networks with unshared parameters. This is equivalent to having one generator for each lighting condition while the attribute code acts as a switch in selecting the respective output to attribute condition for inverse generator or discriminator. The network architecture is illustrated in Fig.~\ref{fig:cycleGAN_unshared}. Note that this is equivalent to having one generator for each lighting condition and therefore each CycleGAN can be trained independently if we further assume unshared networks for discriminator and inverse generator. We conduct experiments on {\accgan} with unshared parameters for all generators and discriminators and report the car model recognition accuracy in Table~\ref{tab:recog_supp}. We observe some improvement in recognition accuracy with unshared models; for example, M7 or M14 with unshared parameters achieve higher accuracy than the ones with shared parameters. On the other hand, M9 with unshared parameters performs a bit lower than the one with shared parameters.

We visualize in Fig.~\ref{fig:{\accgan}_visual} and \ref{fig:{\accgan}_visual_yaw} the photometric transformed images by {\accgan} in both versions of shared and unshared parameters. Besides slight performance improvement for {\accgan} with unshared parameters, we do not observe significant qualitative difference comparing to {\accgan} with shared parameters. Eventually, we believe that the model with shared parameters is more promising for further investigation considering the expansibility of the methods with large number of attribute configurations as well as other interesting properties such as continuous interpolation between attribute configurations.

\newpage
\begin{figure*}[htbp]
\begin{center}
\includegraphics[width=0.45\textwidth]{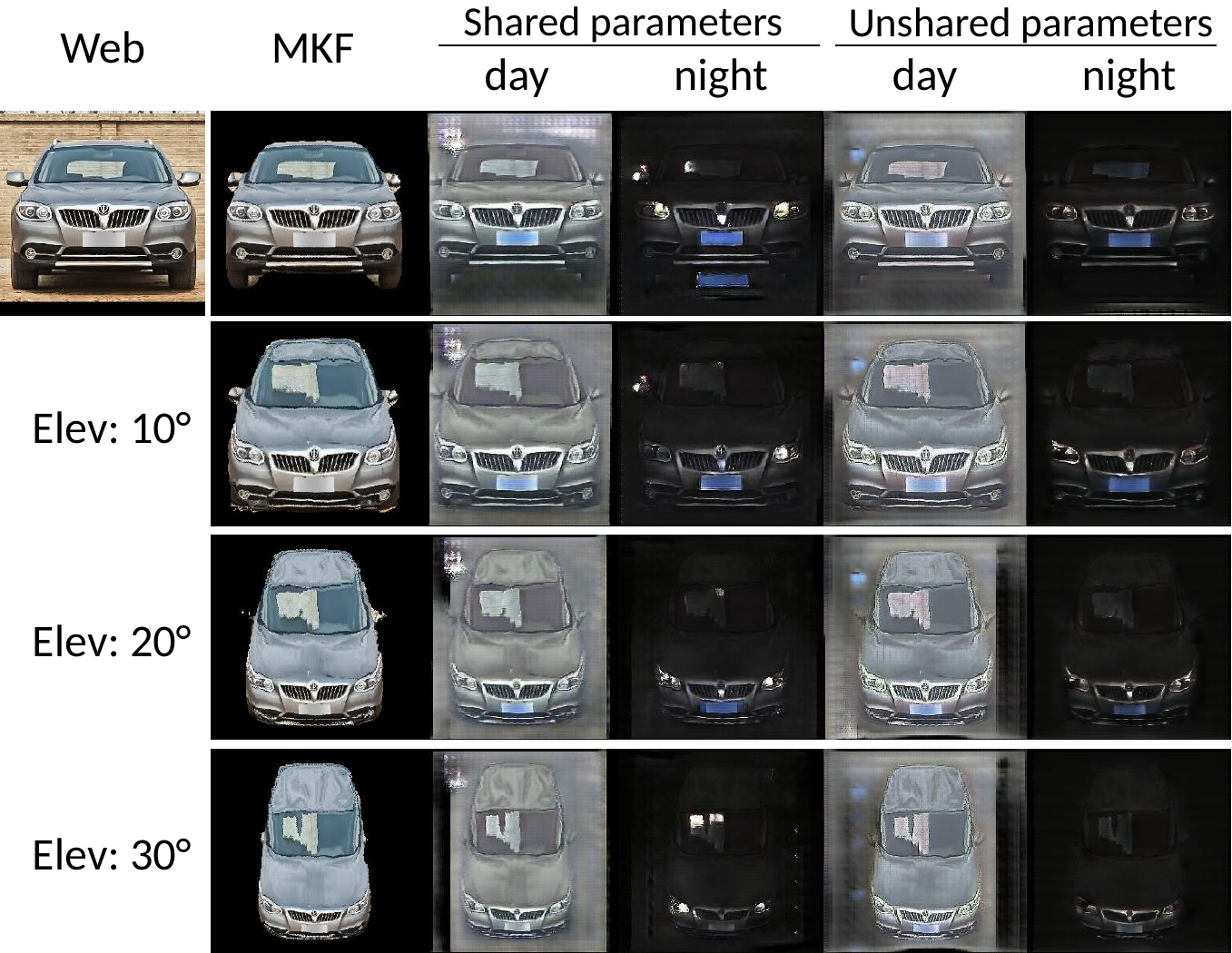}\hspace{0.08in}
\includegraphics[width=0.45\textwidth]{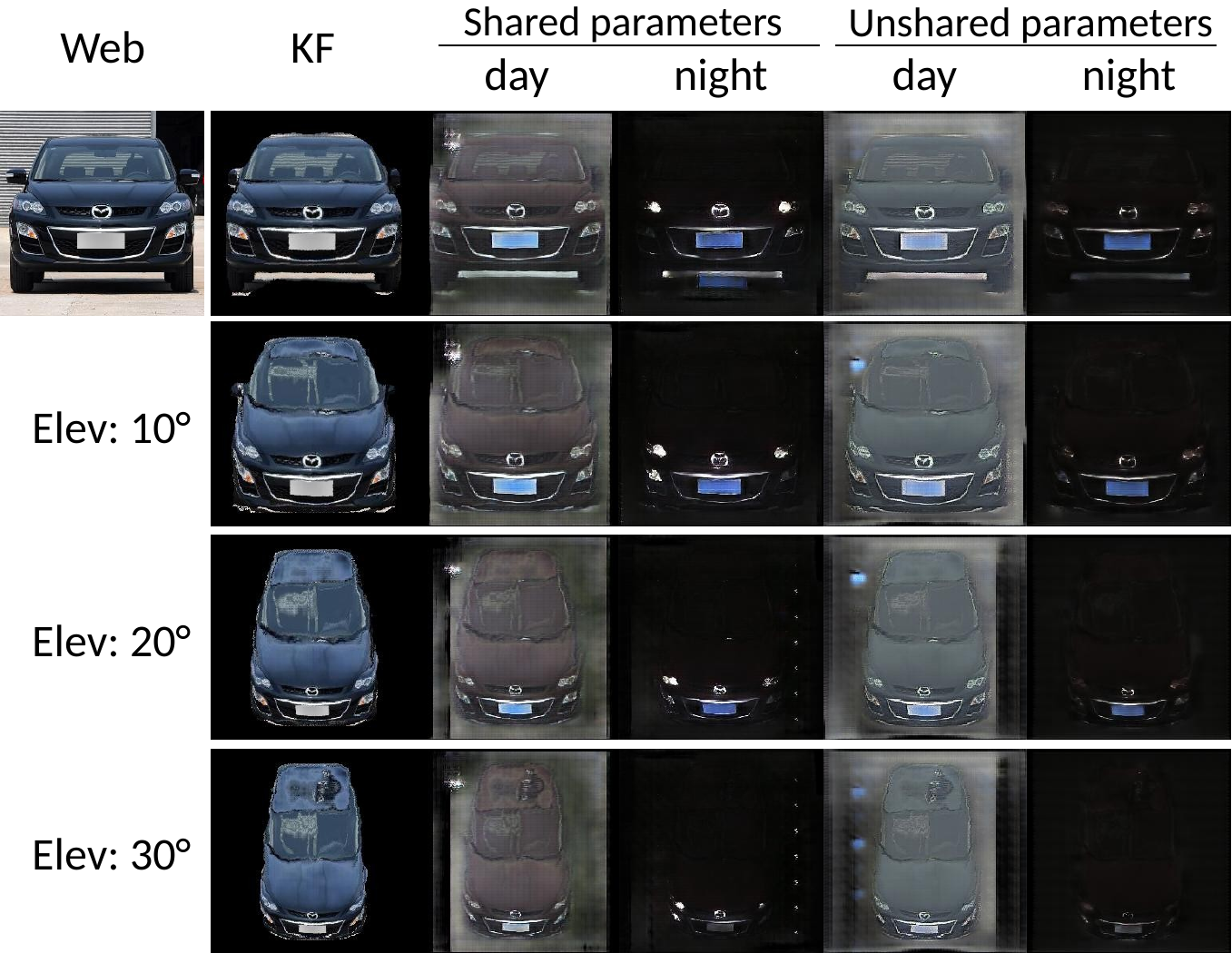}
\vspace{0.05in}
\includegraphics[width=0.45\textwidth]{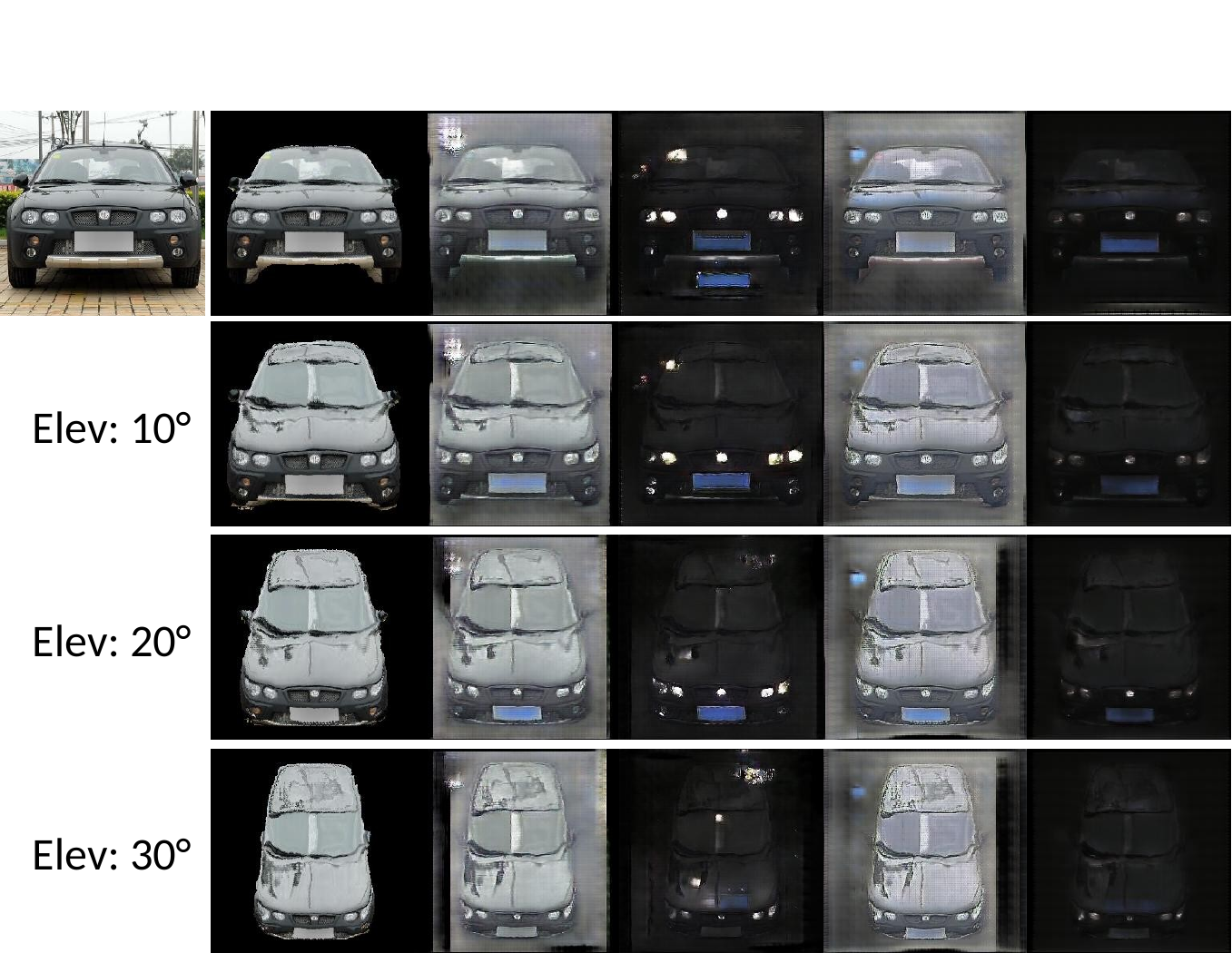}\hspace{0.08in}
\includegraphics[width=0.45\textwidth]{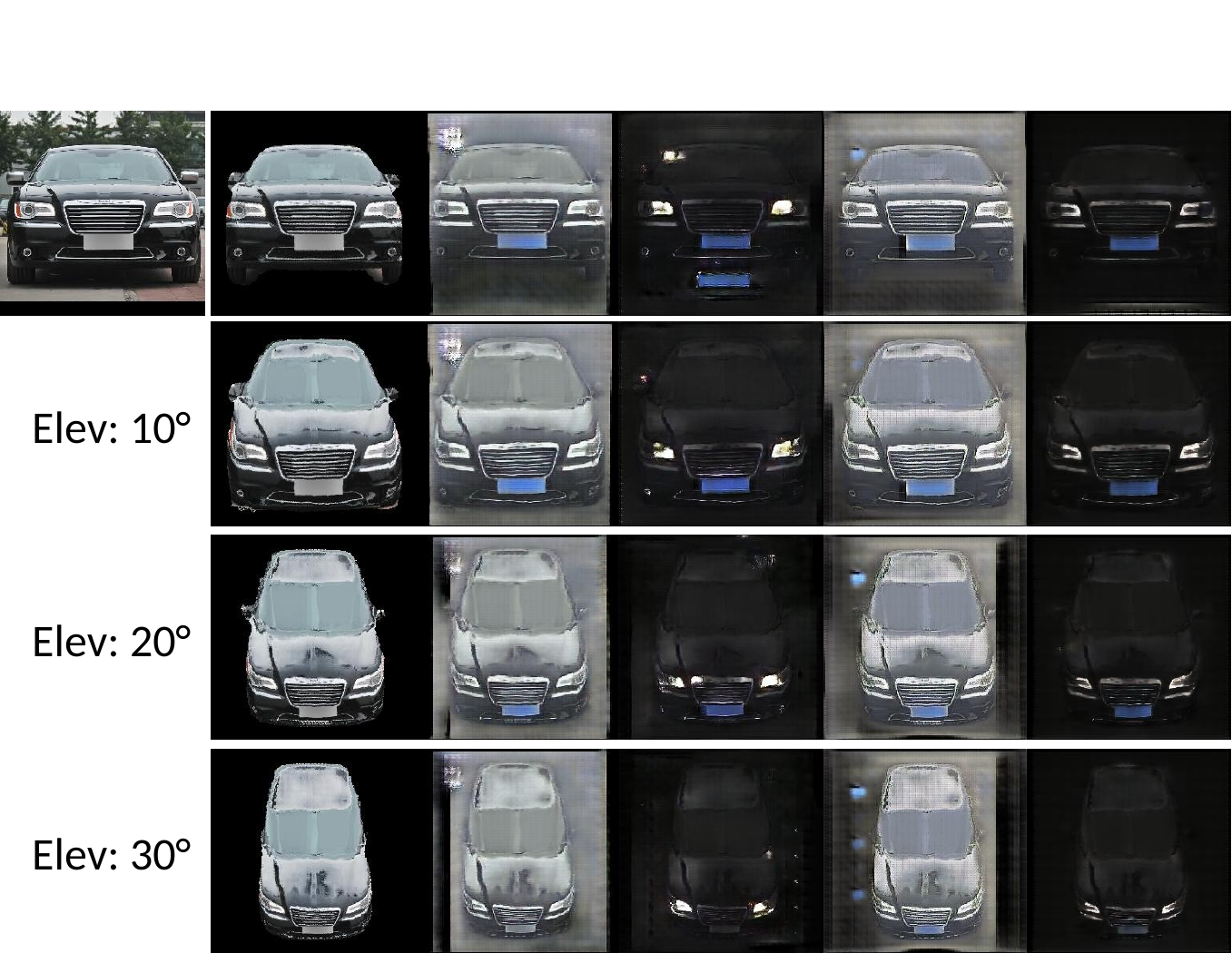}
\vspace{0.05in}
\includegraphics[width=0.45\textwidth]{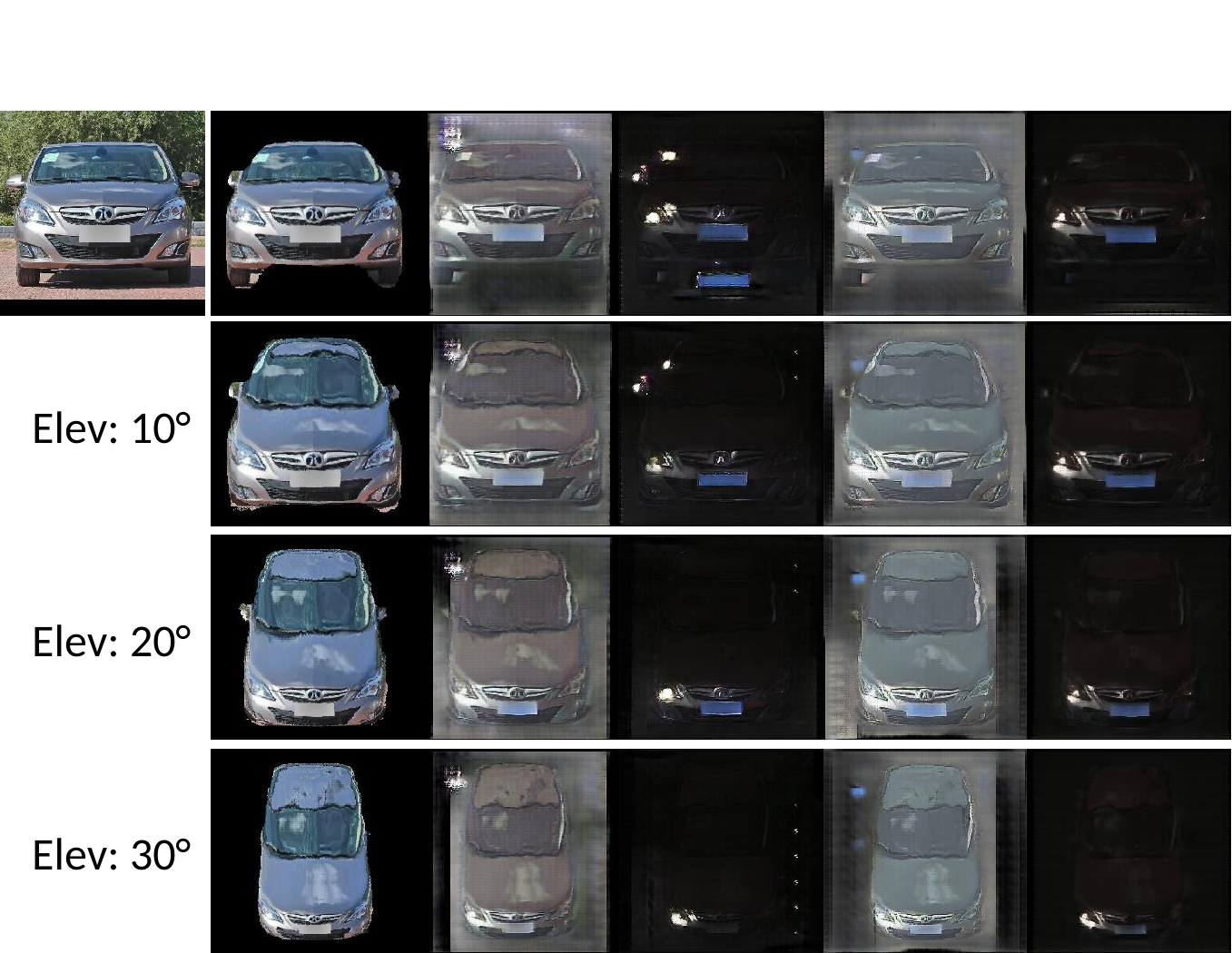}\hspace{0.08in}
\includegraphics[width=0.45\textwidth]{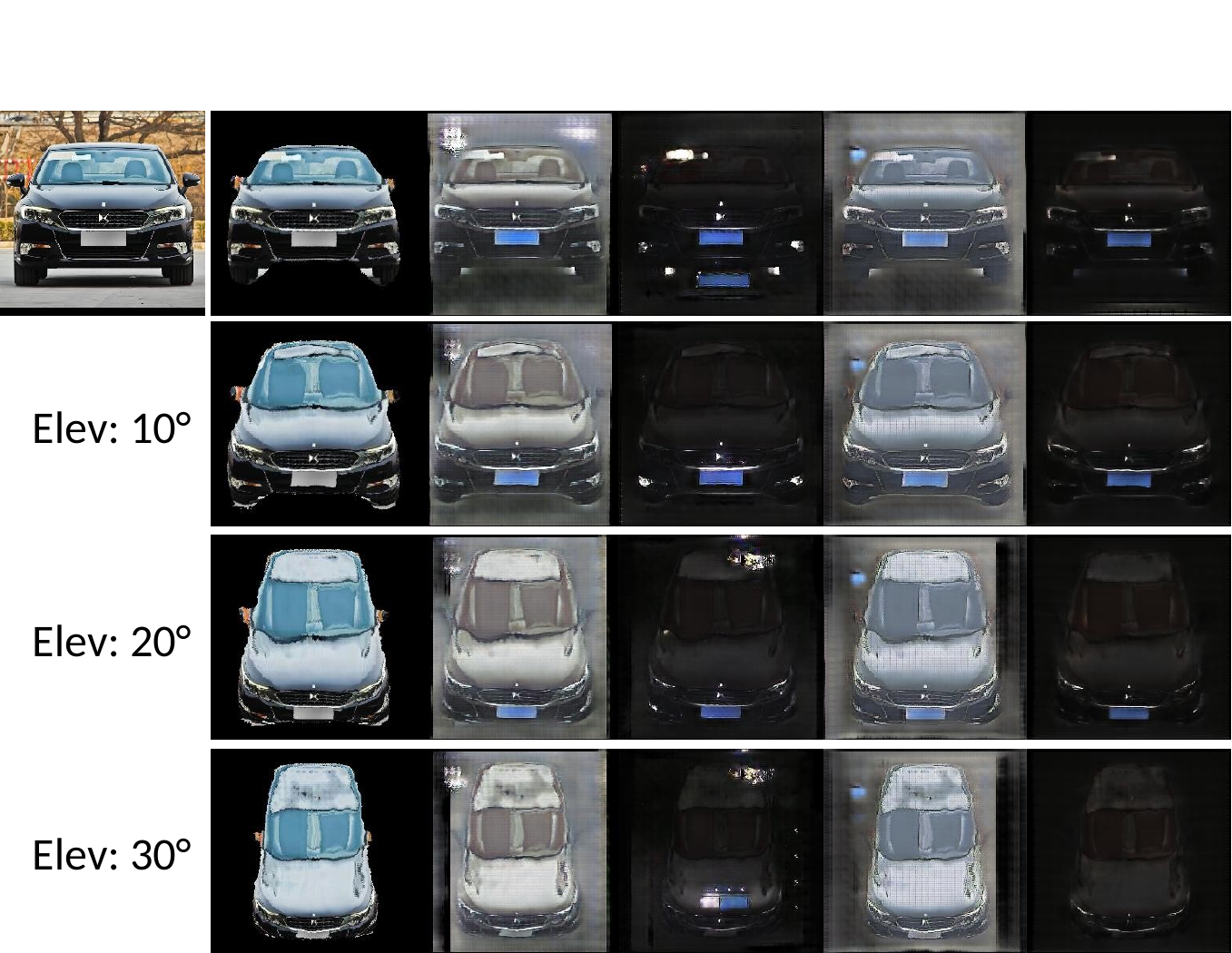}
\caption{Visualization of synthesized images by photometric transformations using {\accgan}s with shared and unshared parameters.\label{fig:{\accgan}_visual}}
\end{center}
\end{figure*}

\begin{figure*}[htbp]
\begin{center}
\includegraphics[width=0.45\textwidth]{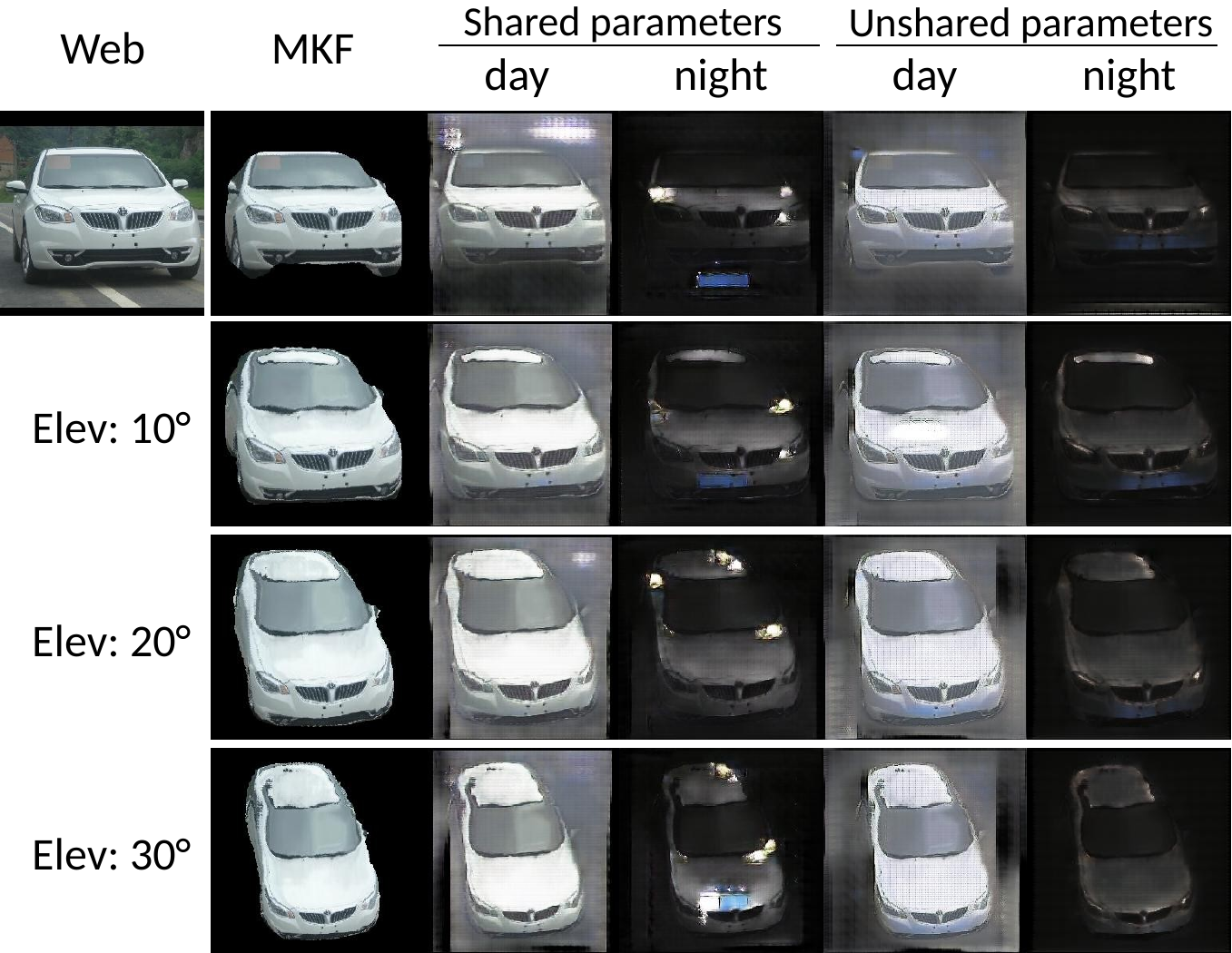}\hspace{0.08in}
\includegraphics[width=0.45\textwidth]{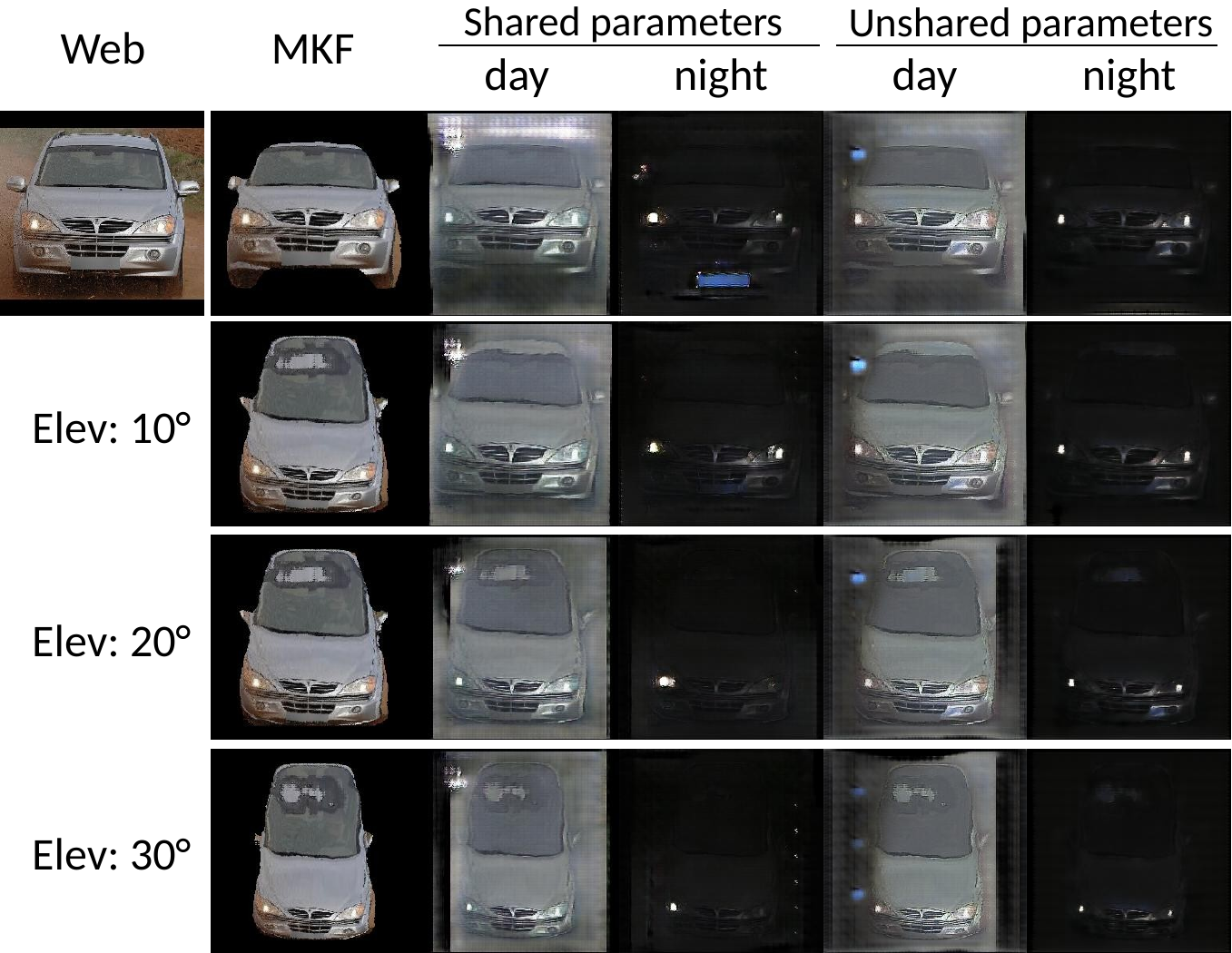}
\vspace{0.05in}
\includegraphics[width=0.45\textwidth]{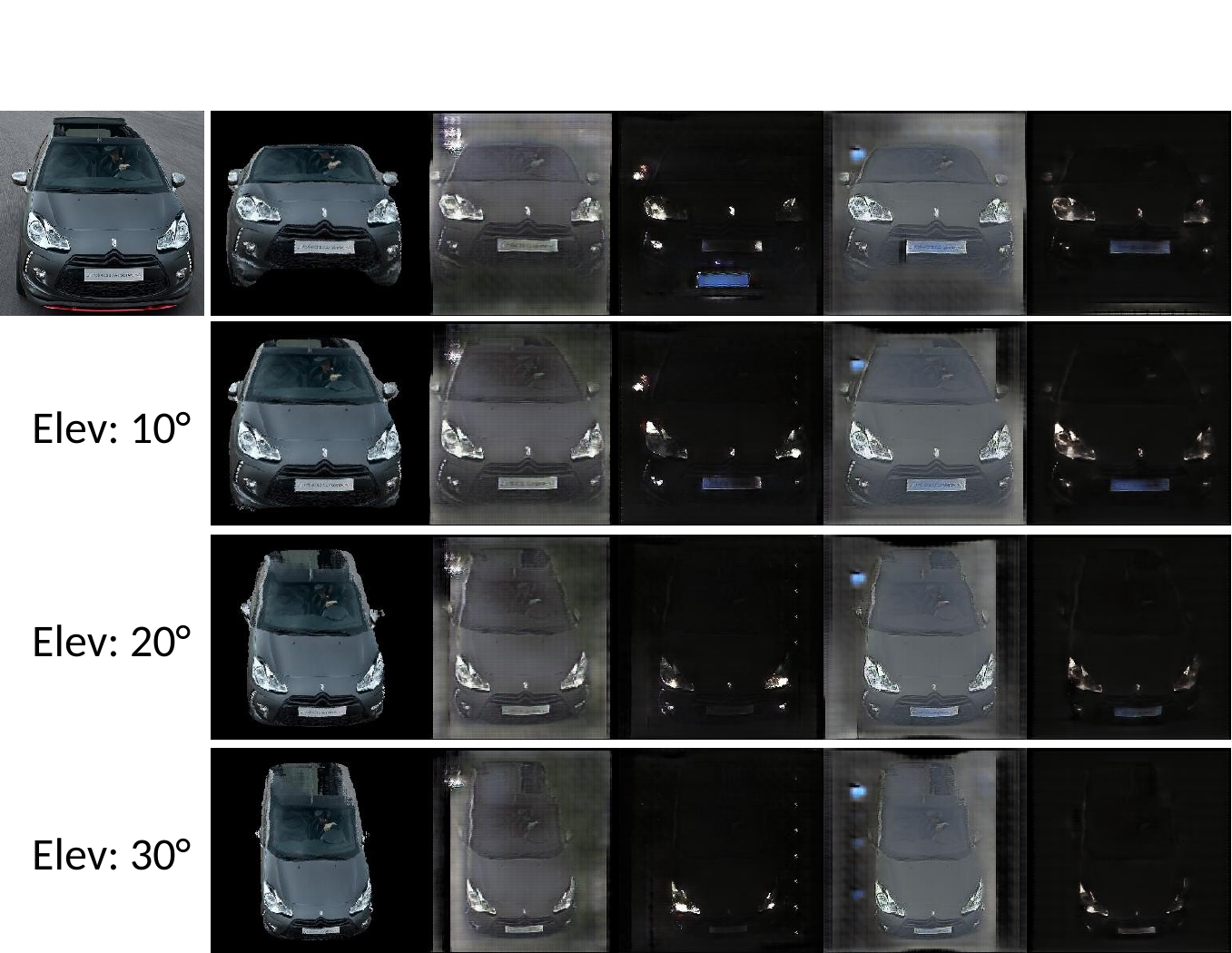}\hspace{0.08in}
\includegraphics[width=0.45\textwidth]{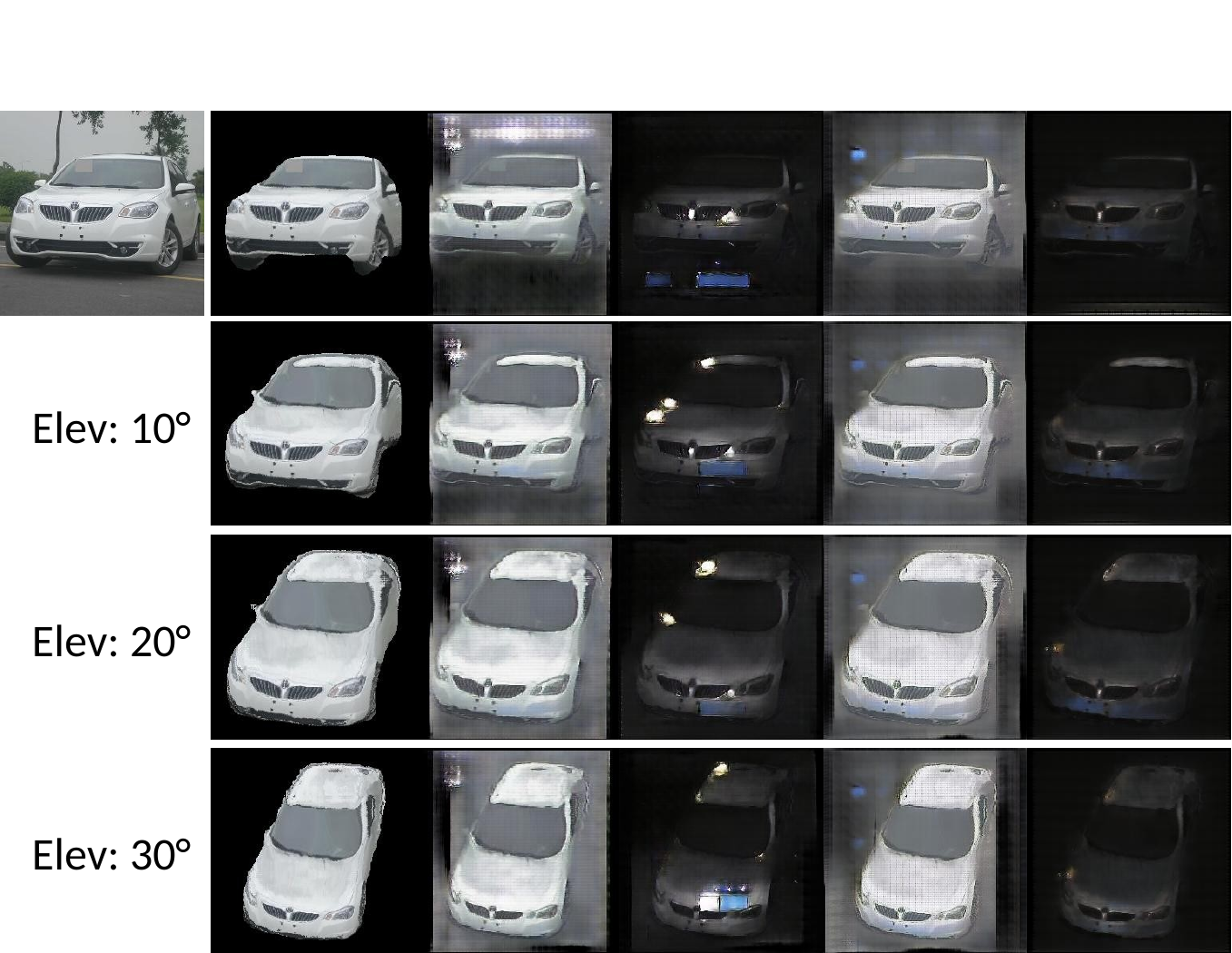}\\
\caption{Visualization of synthesized images by photometric transformations using {\accgan}s with shared and unshared parameters for web images with different yaw angles from $0^{\circ}$.\label{fig:{\accgan}_visual_yaw}}
\end{center}
\end{figure*}

\newpage

\end{document}